\documentclass[letterpaper]{article}
\usepackage{uai2018}
\usepackage[margin=1in]{geometry}
\usepackage{times}

\usepackage{url}
\usepackage{cite}
\PassOptionsToPackage{hyphens}{url}
\newif\ifdsfonts
\dsfontstrue
\usepackage{dsfont}
\usepackage{header,mathtools}
\usepackage{amsmath,amsthm,amssymb}
\DeclareMathOperator{\argmax}{\mathrm{argmax}}
\newcommand{\spaceit}{$\ \ \ $}
\newcommand{\class}{\xi}
\newcommand{\p}{p}
\usepackage{multirow}
\usepackage{xr}
 
\usepackage[round]{natbib}
\let\cite\citep

\newcommand*\samethanks[1][\value{footnote}]{\footnotemark[#1]}

\begin{document} 

\title{Combinatorial Bandits for Incentivizing Agents with Dynamic Preferences}

\author{Tanner Fiez\thanks{\ \ Authors contributed equally.}, \ Shreyas Sekar\samethanks, \ Liyuan Zheng, \normalfont{and} \bf{Lillian J. Ratliff} \\
Electrical Engineering Department, University of Washington\\
}

\maketitle

\begin{abstract}
The design of personalized incentives or recommendations to improve user engagement is gaining prominence as digital platform providers continually emerge. We propose a multi-armed bandit
framework for matching incentives to users,
whose preferences are unknown \emph{a priori} and evolving
dynamically in time, in a resource constrained environment. 
We design an algorithm that combines ideas from three distinct domains: (i) a greedy matching paradigm, (ii) the upper confidence
bound algorithm (UCB) for bandits, and (iii) mixing times from the theory of Markov chains.
For this algorithm, we 
provide theoretical bounds on the regret and demonstrate its performance via both synthetic
and realistic (matching supply and demand in a bike-sharing platform) examples.
\end{abstract}

\thispagestyle{empty}

\section{INTRODUCTION}
\label{sec:intro}
The theory of
\emph{multi-armed bandits} plays a key role in enabling
personalization in the digital economy~\cite{scott2015multi}. Algorithms from this domain have
successfully been deployed in a diverse array of applications including online
advertising~\cite{mehta2011online,lu2010contextual}, crowdsourcing~\cite{ThanhSRJ14}, 
content recommendation~\cite{liCLS10}, and selecting user-specific incentives~\cite{ghoshH13,jainNN14} (e.g., a retailer offering discounts).  On the theoretical side,
this has been complemented by a litany of \emph{near-optimal regret bounds} for
multi-armed bandit settings with rich combinatorial structures and complex agent behavior models~\cite{chenWYW16,gai:2011aa,kveton2015tight,saniLM12}. At a high level, the broad appeal of bandit
approaches for allocating resources to human agents stems from its focus on
balancing \emph{exploration} with \emph{exploitation}, thereby allowing a
 decision-maker to efficiently identify users' preferences without sacrificing short-term rewards.

Implicit in most of these works is the notion that in large-scale environments, a designer can simultaneously allocate resources to multiple users by running independent bandit instances. In reality, such independent
decompositions do not make sense in applications where  
resources are subject to physical or monetary constraints. In simple terms, matching an agent to a resource immediately constrains the set of resources to which another agent can be matched. Such supply constraints may arise even when dealing with intangible products. For instance, social media platforms (e.g., Quora)~seek to maximize user participation by offering incentives in the form of increased recognition---e.g., featured posts or badges~\cite{immorlicaSS15}. Of course, there are supply constraints on the number of posts or users that can be featured at a given time. As a consequence of these coupling constraints, much of the existing work on multi-armed bandits does not extend naturally to multi-agent economies.

Yet, another important aspect 
not addressed by the literature concerns human behavior. Specifically, users'
preferences over the various resources may be dynamic---i.e.~evolve in time as they are repeatedly exposed to the available options. 
The problem faced by a designer in such a dynamic environment is compounded by the lack of information regarding each user's current state or beliefs, as well as how these beliefs influence their preferences and their evolution in time. 

Bearing in mind these limitations, we study a multi-armed bandit problem for
matching multiple agents to a finite set of incentives\footnote{We use the term
incentive broadly to refer to any resource or action available to the agent.
That is, incentives are not limited to monetary or financial mechanisms.}: each
incentive belongs to a category and global capacity constraints control the
number of incentives that can be chosen from each category. In our model, each
agent has a \emph{preference profile} or a \emph{type} that determines its rewards for being
matched to different incentives. The agent's type evolves according to a Markov
decision process (MDP), and therefore, the rewards vary over time \emph{in a correlated fashion}. 

Our work is primarily motivated by the problem faced by a technological platform
that seeks to not just maximize user engagement but also to encourage users to make
changes in their \emph{status quo} decision-making process by offering
incentives. For concreteness, consider a bike-sharing service---an application
we explore in our simulations---that seeks to identify optimal incentives for each user from a finite bundle of options---e.g., varying discount levels, free future rides, bulk ride offers, etc. Users' preferences over the incentives may evolve with time depending on their current type, which in turn depends on their previous experience with the incentives. In addition to their marketing benefits, such incentives can serve as a powerful instrument for \emph{nudging} users to park their bikes at alternative locations---this can lead to spatially balanced supply and consequently, lower rejection rates~\cite{singlaSBMMK15}.

\subsection{CONTRIBUTIONS AND ORGANIZATION}
Our objective is to design a multi-armed bandit algorithm that repeatedly
matches agents to incentives in order to minimize the cumulative \emph{regret} over
a finite time horizon. Here, regret is defined as the difference in the reward
obtained by a problem specific benchmark strategy and the proposed algorithm (see Definition \ref{defn:regret}). A
preliminary impediment in achieving this goal is the fact that the capacitated
matching problem studied in this work is NP-Hard even in the offline case. The major challenge therefore is whether \emph{we can achieve sub-linear (in the length of the horizon) regret in the more general  matching environment without any information on the users' underlying beliefs or how they evolve}?

Following preliminaries (Section~\ref{sec:formulation}), we introduce a simple greedy algorithm that provides
a $1/3$--approximation to the optimal offline matching solution
(Section~\ref{sec:offline}).
Leveraging this first contribution, the central result in this paper (Section~\ref{sec:bandit}) is a new multi-armed bandit
algorithm---\emph{MatchGreedy-EpochUCB} (MG-EUCB)---for
capacitated matching problems with time-evolving rewards. Our algorithm obtains
logarithmic (and hence sub-linear) regret even for this more general bandit
problem. The proposed approach combines ideas from three distinct domains: (i)
the $1/3$--rd approximate greedy matching algorithm, (ii) the traditional UCB algorithm~\cite{auer:2002aa}, and (iii) mixing times from the theory of Markov chains. 

We validate our theoretical results (Section~\ref{sec:experiments}) by
performing simulations on both synthetic and realistic instances
derived using data obtained from a Boston-based bike-sharing service
\emph{Hubway}~\cite{hubway}. 
We compare our algorithm to existing UCB-based approaches and show that the proposed method enjoys favorable convergence rates, computational efficiency on large data sets, and does not get stuck at sub-optimal matching solutions.

\subsection{BACKGROUND AND RELATED WORK} 
Two distinct features separate our model from the majority of work on 
the multi-armed bandit problem: (i) our focus on a capacitated
matching problem with finite supply (every user cannot be matched to their
optimal incentive), and (ii) the rewards associated with each agent evolve in a
correlated fashion but the designer is unaware of each agent's current state.
Our work is closest to \cite{gai:2011aa} which considers a matching problem with Markovian rewards. However, in their model the rewards associated with each edge evolve independently of the other edges; as we show via a simple example in Section~\ref{sec:challenges}, the correlated nature of rewards in our instance can lead to additional challenges and convergence to sub-optimal matchings if we employ a traditional approach as in~\cite{gai:2011aa}.

Our work also bears conceptual similarities to the rich literature on
combinatorial bandits~\cite{badanidiyuruKS13,chenWYW16,kvetonWAEE14, kveton2015tight, wen2015efficient}.
However, unlike our work, these papers consider a model where the distribution
of the rewards is static in time. For this reason, efficient learning algorithms leveraging oracles to solve generic constrained combinatorial optimization problems developed for the combinatorial semi-bandit setting~\cite{chenWYW16, kveton2015tight} face similar limitations in our model as the approach of~\cite{gai:2011aa}. Moreover, the rewards in our problem may not have a linear structure so the approach of~\cite{wen2015efficient} is not applicable.

The novelty in this work is not the combinatorial aspect but the interplay between combinatorial bandits and the edge rewards evolving according to an MDP. When an arm is selected by an oracle, the reward of every edge in the graph evolves—---how it evolves depends on which arm is chosen. If the change occurs in a sub-optimal direction, this can affect future rewards. Indeed, the difficulties in our proofs do not stem from applying an oracle for combinatorial optimization, but from bounding the secondary regret that arises when rewards evolve in a sub-optimal way.

Finally, there is a  somewhat parallel body of work on single-agent
reinforcement learning techniques~\cite{jaksch:2010aa,
mazumdar:2017aa,azar:2013aa,ratliff:2018aa} and expert selection where the rewards on the arms evolve in a correlated fashion as in our work. In addition to our focus on multi-agent matchings, we remark that many of these works assume that the designer is aware (at least partially) of the agent's exact state and thus, can eventually infer the nature of the evolution. Consequently, a major contribution of this work is the extension of UCB-based approaches to solve MDPs with a \emph{fully unobserved state} and rewards associated with each edge that evolve in a correlated fashion.

\ifdsfonts

\section{PRELIMINARIES}
\label{sec:formulation}
A designer faces the problem of matching $m$ agents to incentives
(more generally jobs, goods, content, etc.)  without violating certain capacity
constraints.
We model this setting by means of a bipartite graph $(\mathcal{A}, \mathcal{I},
\mathcal{P})$ where $\mc{A}$ is the set of agents, $\mc{I}$ is the
set of incentives to which the agents can be matched, and
 $\mathcal{P} = \mathcal{A} \times \mathcal{I}$ is the set of all
pairings between agents and incentives. We sometimes refer to $\mathcal{P}$ as
the set of arms.
In this regard, a matching is a set $M \subseteq \mathcal{P}$ such that every agent $a \in \mc{A}$ and incentive $i \in \mc{I}$ is present in at most one edge belonging to $M$.

Each agent $a \in \mc{A}$ is associated with a type or state $\theta_a \in \Theta_a$, which
influences the reward received by this agent when matched with some incentive $i
\in \mathcal{I}$. When agent $a$ is matched to incentive $i$, its type evolves according to a Markov process with transition probability
kernel $P_{a,i}: \Theta_a \times \Theta_a \rightarrow [0,1]$.
Each pairing or edge of the bipartite graph is associated with some reward that
depends on the agent--incentive pair, $(a,i)$, as well as the type $\theta_a$. 

The designer's policy (algorithm) is to compute a matching repeatedly over a
finite time horizon in order to maximize the expected aggregate reward. In this
work, we restrict our attention to a specific type of multi-armed bandit
algorithm that we refer to as an \emph{epoch mixing policy}. Formally, the
execution of such a policy $\alpha$ is divided into a finite number of time
indices $[n] = \{1, 2, \ldots, n\}$, where $n$ is the length of the time
horizon. In each time index $k \in [n]$, the policy selects a matching
$\alpha(k)$ and repeatedly `plays' this matching for $\tau_k > 0$ iterations
within this time index.  We refer to the set of iterations within a time index
collectively as an \emph{epoch}. That is, within the $k$--{th} epoch, for each edge $(a,i) \in \alpha(k)$, agent $a$ is matched to incentive $i$ and the agent's type is allowed to evolve for $\tau_k$ iterations. 
In short, an epoch mixing policy proceeds in two time scales---each selection of a matching corresponds to an epoch comprising of $\tau_k$ iterations for $k \in [n]$, and there are a total of $n$ epochs. 
It is worth noting that an epoch-based policy was used in the UCB2 algorithm~\cite{auer:2002aa}, albeit with stationary rewards.

Agents' types evolve based on the incentives to which they are matched. Suppose
that $\beta^{(k)}_a$ denotes the type distribution on $\Theta_a$ at epoch $k$ 
and $i \in \mc{I}$ is the
incentive to which agent $a$ is matched by $\alpha$ (i.e., $(a,i) \in
\alpha(k)$). Then,  $\beta^{(k+1)}_a(\theta_a)=\sum_{\theta'\in \Theta_a} P_{a,i}^{\tau_k}(\theta',\theta_a)\beta^{(k)}_a(\theta').$

For epoch $k$, the rewards are averaged over the $\tau_k$ iterations in that
epoch. Let $r^{\theta}_{a,i}$ denote the reward received by agent $a$ when it is
matched to incentive $i$ given type $\theta\in \Theta_a$. We assume that  $r^{\theta}_{a,i} \in [0,1]$ and is drawn from a distribution $\mc{T}_r(a,i,\theta)$. The reward distributions for different edges and states in $\Theta_a$ are assumed to be independent of each other. 
Suppose that an algorithm $\alpha$ selects the edge $(a,i)$ for $\tau$
iterations within an epoch. The observed reward at the end of this epoch is
taken to be the time-averaged reward over the epoch. Specifically, suppose that the $k$--th epoch proceeds for $\tau_k$ iterations
beginning with time $t_k$---i.e. one plus the total iterations completed before
this---and ending at time $t_{k+1} - 1 = t_k + \tau_k - 1$,
and that $\theta_a(t)$ denotes agent $a$'s state at time $t \in [t_k, t_{k+1} - 1]$. Then, the time-averaged reward in the epoch is given by 
$\textstyle\bs{r}^{\theta_a(t_k)}_{a,i}=\frac{1}{\taua{k}}\sum_{t=t_k}^{t_{k+1}-1}r_{a,i}^{\theta_a(t)}.$
We use the state as a superscript to denote dependence of the reward on the agent's type at the beginning of the epoch. Finally, the total (time-averaged) reward due to a matching $\alpha(k)$ at the end of an epoch can be written as $\sum_{(a,i) \in \alpha(k)} \bs{r}^{\theta_a(t_k)}_{a,i}.$ 

We assume that the Markov chain corresponding to each edge $(a,i) \in \mc{P}$ is
\emph{aperiodic} and \emph{irreducible}~\cite{levin:2009aa}. We denote the stationary or steady-state distribution for this edge as $\pi_{a,i}: \Theta_a \rar [0,1]$. 
Hence, we define the expected reward for edge $(a,i)$, given its stationary
distribution, to be $\textstyle\mu_{a,i}=\mb{E}\left[\sum_{\theta \in \Theta_a}
r^{\theta}_{a,i}\pi_{a,i}(\theta)\right]$
where the expectation is with respect to the distribution on the reward given
$\theta$.

\subsection{CAPACITATED MATCHING}
\label{sec:matching}
Given $\mathcal{P} = \mathcal{A} \times \mathcal{I}$, the designer's goal at the
beginning of each epoch is to select a matching $M \subseteq \mathcal{P}$---i.e.~a
collection of edges---that satisfies some cardinality constraints.
We  partition the edges in
$\mc{P}$ into a mutually exclusive set of classes
allowing for edges possessing identical
characteristics to be grouped together. In the bike-sharing example, the various classes could denote types of incentives---e.g., edges that match agents to discounts, free-rides, etc.
 Suppose that  $\mathcal{C} = \{\class_1, \class_2, \ldots, \class_q\}$ denotes a
partitioning of the edge set such that (i) $\class_j \subseteq \mc{P}$ for all $1
\leq j \leq q$, (ii) $\bigcup_{j=1}^q \class_j = \mc{P}$, and (iii) $\class_j \cap
\class_{j'} = \emptyset$ for all $j \neq j'$. We refer to each $\class_j$ as a class and for any given edge $(a,i) \in \mc{P}$, use $c(a,i)$ to denote the class that this edge belongs to, i.e., $(a,i) \in c(a,i)$ and $c(a,i) \in \mc{C}$.

Given a capacity vector $\boldsymbol{b} = (b_{\class_1}, \ldots, b_{\class_q})$ indexed on the set of classes, we say that a matching $M \subseteq \mc{P}$ is a feasible solution to the capacitated matching problem if:
\begin{itemize}[itemsep=-5pt,topsep=-5pt, leftmargin=15pt]
    \item[a)] for every $a \in \mc{A}$ (resp., $i \in \mc{I}$), the
        matching must contain at most one edge containing this agent (resp., incentive)

    \item[b)] and, the total number of edges from each class $\class_j$
        contained in the matching cannot be larger than $b_{\class_j}$.
\end{itemize}

In summary, the \emph{capacitated matching problem} can be formulated as the following integer program:
\begin{equation}
\begin{aligned}
\quad \text{max} & \quad\textstyle \sum_{(a,i) \in \mathcal{P}} w(a,i) x(a,i)  \\
\text{s.t.} & \quad\textstyle \sum_{i \in \mathcal{I}}x(a,i) \leq 1 \quad \forall a \in \mathcal{A}
\\
& \quad\textstyle \sum_{a \in \mathcal{A}}x(a,i) \leq 1 \quad \forall i \in \mathcal{I}
\\
& \quad\textstyle \sum_{(a,i) \in \class_j}x(a,i) \leq b_{\class_j}, \quad \forall \class_j \in \mathcal{C} \\
& \quad x(a,i) \in \{0,1\}, \quad \forall (a,i) \in \mathcal{P}
\end{aligned}\tag{P1}
\label{eq:lp_matching}
\end{equation}
We use the notation 
$\{ \mc{P}, \mc{C}, \bs{b},  (w(a,i))_{(a,i) \in \mc{P}}\}$
for a \emph{capacitated matching problem instance}.
In~\eqref{eq:lp_matching}, $w(a,i)$ refers to the weight or the reward to be
obtained from the given edge. The term $x(a, i)$ is an indicator on whether the edge $(a, i)$ is included in the solution to~\eqref{eq:lp_matching}. Clearly, the goal is to select a maximum weight matching subject to the constraints. In our online bandit problem, the
designer's actual goal in a fixed epoch $k$ is to maximize the quantity $\sum_{(a,i)
    \in \mc{P}} \bs{r}^{\theta_a(t_k)}_{a,i} x(a,i)$, i.e., $w(a,i) =
    \bs{r}^{\theta_a(t_k)}_{a,i}$. However, since the reward distributions and the
    current user type are not known beforehand, our MG-EUCB algorithm (detailed in Section~\ref{sec:ucb}) approximates this objective by setting the weights to be the average observed reward from the edges in combination with the corresponding confidence bounds.

\subsection{TECHNICAL CHALLENGES}
\label{sec:challenges}
There are two key obstacles involved in extending traditional bandit approaches to our combinatorial setting with evolving rewards, namely, \emph{cascading
	sub-optimality} and \emph{correlated convergence}. The first phenomenon occurs when an agent $a$ is matched to a sub-optimal arm $i$ (incentive) because its optimal arm $i^*$ has already been assigned to another agent. Such sub-optimal pairings have the potential to cascade, e.g., when another agent $a_1$ who is matched to $i$ in the optimal solution can no longer receive this incentive and so on. Therefore, unlike the classical bandit analysis, the selection of sub-optimal arms cannot be directly mapped to the empirical rewards.


\emph{Correlated Convergence}. As mentioned previously, in our model, the rewards
depend on the type or state of an agent, and hence, the reward distribution on any given edge $(a,i)$ can vary even when the algorithm does not select this edge. As a
result, a na\"{i}ve application of a bandit algorithm can severely under-estimate
the expected reward on each edge and eventually converge to a sub-optimal
matching. A concrete example of the poor convergence effect is provided in
Example~\ref{ex:cuteex}. In Section~\ref{sec:ucb}, we describe how our central bandit algorithm limits the damage due to cascading while simultaneously avoiding the correlated convergence problem.

%
%

\begin{example}[Failure of Classical UCB]
    \label{ex:cuteex}

Consider a problem instance with two agents $\mc{A} = \{a_1, a_2\}$, two incentives $\mc{I}=\{i_1,i_2\}$ and identical state space i.e.,  $\Theta_{a_1}= \Theta_{a_2} = \{\theta_1,
\theta_2\}$. The transition matrices and deterministic rewards for the agents for being matched to each incentive are depicted pictorially below: we assume that $\epsilon > 0$ is a sufficiently small constant. 

\begin{figure}[h]
	\centering
                   \begin{tikzpicture}[font=\sffamily, scale=0.74, every node/.style={transform shape}]
	
	\node[state,
	text=yellow,
	draw=none,
	fill=gray!50!black] (s) {$\theta_1$};
	\node[state,
	right=1.5cm of s,
	text=blue!30!white, 
	draw=none, 
	fill=gray!50!black] (r) {$\theta_2$};
	
    \node[below=0.0cm  of s] (som) {$r^{\theta_1}_{a_1,i_1}=0$}; 
		\node[below right=0.1cm and -0.75cm of r] 
        (som) {$r^{\theta_2}_{a_1,i_1}=1$}; 
	
	\draw[every loop,
	auto=right,
	line width=0.5mm,
	>=latex,
	draw=orange,
	fill=orange]
	(s) edge[bend right, auto=left]  node {1} (r)
	(r) edge[bend right, auto=right] node {$\epsilon$} (s)
    (s) edge[loop above]             node[left=0.1cm] {0} (s)
    (r) edge[loop above ]             node[left=0.1cm] {$1-\epsilon$} (r);
    \node[left=0.25cm of s,text width = 8mm] {Edge $(a_1, i_1)$};
    \node[state,
text=yellow,
draw=none,
fill=gray!50!black, below=1.5cm of s] (s1) {$\theta_1$};
\node[state,
right=1.5cm of s1,
text=blue!30!white, 
draw=none, 
fill=gray!50!black] (r1) {$\theta_2$};
\node[below =0.0cm of s1] (som1) {$r^{\theta_1}_{a_1,i_2}=0.5$}; 
\node[below=0.0cm of  r1] (som2) {$r^{\theta_2}_{a_1,i_2}=0.5$}; 

\draw[every loop,
auto=right,
line width=0.5mm,
>=latex,
draw=orange,
fill=orange]
(s1) edge[bend right, auto=left]  node {$\epsilon$} (r1)
(r1) edge[bend right, auto=right] node {1} (s1)
(s1) edge[loop above]             node[left=0.1cm] {$1-\epsilon$} (s1)
(r1) edge[loop above]             node[left=0.1cm] {0} (r1);
\node[left=0.25cm of s1,text width = 8mm] {Edge $(a_1, i_2)$};

	\node[state,
	text=yellow,
	right=4.5cm of s,
	draw=none,
	fill=gray!50!black] (s2) {$\theta_1$};
	\node[state,
	right=1.5cm of s2,
	text=blue!30!white, 
	draw=none, 
	fill=gray!50!black] (r2) {$\theta_2$};
	
	\node[below=0.0cm  of s2] (som22) {$r^{\theta_1}_{a_2,i_1}=0.5$}; 
	\node[below right=0.1cm and -1.25cm of r2] 
	(som23) {$r^{\theta_2}_{a_2,i_1}=0.5$}; 
	
	\draw[every loop,
	auto=right,
	line width=0.5mm,
	>=latex,
	draw=orange,
	fill=orange]
	(s2) edge[bend right, auto=left]  node {$\epsilon$} (r2)
	(r2) edge[bend right, auto=right] node {$1$} (s2)
	(s2) edge[loop above]             node[left=0.1cm] {$1-\epsilon$} (s2)
	(r2) edge[loop above ]             node[left=0.1cm] {$0$} (r2);
	\node[left=0.2cm of s2,text width = 8mm] {Edge $(a_2, i_1)$};
	
	\node[state,
	text=yellow,
	draw=none,
	fill=gray!50!black, below=1.5cm of s2] (s3) {$\theta_1$};
	\node[state,
	right=1.5cm of s3,
	text=blue!30!white, 
	draw=none, 
	fill=gray!50!black] (r3) {$\theta_2$};
	
	\draw[every loop,
	auto=right,
	line width=0.5mm,
	>=latex,
	draw=orange,
	fill=orange]
	(s3) edge[bend right, auto=left]  node {$1$} (r3)
	(r3) edge[bend right, auto=right] node {$\epsilon$} (s3)
	(s3) edge[loop above]             node[left=0.1cm] {$0$} (s3)
	(r3) edge[loop above]             node[left=0.1cm] {$1-\epsilon$} (r3);
	\node[left=0.2cm of s3,text width = 8mm] {Edge $(a_2, i_2)$};

\node[below =0.0cm of s3] (som1) {$r^{\theta_1}_{a_2,i_2}=0$}; 
\node[below =0.0cm of  r3] (som2) {$r^{\theta_2}_{a_2, i_2}=1$}; 
	
	\node[above right = .9cm and 0.1 of s](aa){Agent $a_1$};
	\node[above right = .9cm and 0.1 of s2](aa){Agent $a_2$};
	
	\draw [dashed] (3.55,1) -- (3.55,-3.8);
	
\end{tikzpicture} 

\caption{(a) State transition diagram and reward for each edge: note that the
    state is associated with the agent and not the edge. }
 \label{fig:M1}
\end{figure}
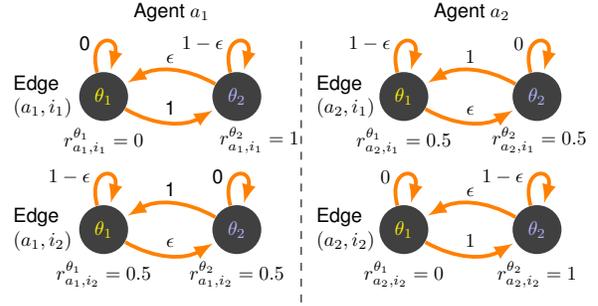
Clearly, the optimal strategy is to repeatedly chose the matching $\{(a_1,i_1),
(a_2,i_2)\}$ achieving a reward of (almost) two in each epoch. 
An implementation of traditional UCB for the
matching problem---e.g., the approach in~\cite{gai:2011aa,chenWYW16, kveton2015tight}---selects a matching based on the
empirical rewards and confidence bounds for a total of $\sum_{k=1}^n \tau_k$
iterations, which are then divided into epochs for convenience. 
This approach converges to the sub-optimal matching of $M=\{(a_1,
i_2), (a_2, i_1)\}$. Indeed, every time the algorithm selects
this matching, both the agents' states are reset to $\theta_1$ and 
when the algorithm explores the optimum matching, the reward consistently
happens to be zero since the agents are in state $\theta_1$. 
Hence, the rewards for the (edges in the) optimum matching are grossly underestimated. 
\end{example}


\section{GREEDY OFFLINE MATCHING}
\label{sec:offline}
In this section, we consider the capacitated matching problem in the offline
case, where the edge weights are provided as input. The techniques developed in
this section serve as a base in order to solve the more general online problem
in the next section.
More specifically, we assume
that we are given an arbitrary instance of the capacitated matching  problem
$\{ \mc{P}, \mc{C}, \bs{b},  (w(a,i))_{(a,i) \in \mc{P}}\} .$ Given this instance, the designer's objective is to
solve~\eqref{eq:lp_matching}. Surprisingly, this problem turns out to be NP-Hard
and thus cannot be optimally solved in polynomial time~\cite{GareyJ79}---this marks a stark contrast with the classic maximum weighted matching problem, which can be solved efficiently using the Hungarian method~\cite{kuhn1955hungarian}.

In view of these computational difficulties, we develop a simple greedy approach
for the capacitated matching problem and formally prove that it results in a
one-third approximation to the optimum solution. The greedy method studied in this work comes with a multitude of desirable
properties that render it suitable for matching problems arising in large-scale
economies. Firstly, the greedy algorithm has a running time of $O(m^2\log m)$,
where $m$ is the number of agents---this near-linear execution time in the number of edges makes it ideal for platforms comprising of a large number of agents. Secondly, since the output of the greedy
algorithm depends only on the ordering of the edge weights and is not sensitive
to their exact numerical value, learning approaches tend to converge faster to
the `optimum solution'. This property is validated by our simulations (see Figure~\ref{fig:ma_optimal}). 
Finally, the performance of the greedy algorithm in practice (e.g., see Figure~\ref{fig:epoch_mean_reward}) appears to be much closer to the
optimum solution than the 1/3 approximation guaranteed by
Theorem~\ref{thm_matchapprox} below.


\begin{algorithm}[t]
	\caption{Capacitated-Greedy Matching Algorithm }
	\label{alg:greedy}
	\begin{algorithmic}[1] 
			\algnewcommand{\AND}{\algorithmicand}
		\Function{MG}{($w(a,i))_{(a,i) \in \mathcal{P}}$, $\bs{b}$}\\
		 $\quad$$G^\ast \gets \emptyset $, $E' \gets \mathcal{P}$\\
		 $\quad$\textbf{while} {$E'  \neq \emptyset$}:\\
		 $\quad\ $ Select $(a, i) = \arg\max_{(a', i') \in E'} w(a',i')$\\
         $\quad \ $\textbf{if}{$|G^\ast \cap c(a,i)| < b_{c(a,i)}$} \textbf{then}\\
			$\quad\ \ \ $ $G^\ast \gets G^\ast \cup \{(a,i)\}$\\
     		$\quad\ \ \ $ $E' \gets E' \setminus \{(a',i')\}$ $\forall (a',i'): a' = a$ or $i'=i$ \label{alggreedy_firstremoval}
            $\quad\ $\textbf{else}\\
			$\quad \ \ \ $ $E' \gets E' \setminus \{(a,i)\}$ \label{alggreedy_secondremoval}\\
			$\quad$\Return $G^\ast$
			\EndFunction
	\end{algorithmic}
\end{algorithm}

\subsection{ANALYSIS OF GREEDY ALGORITHM}
The greedy matching is outlined in Algorithm~\ref{alg:greedy}. Given an instance $ \{ \mc{P}, \mc{C}, \bs{b},  (w(a,i))_{(a,i) \in \mc{P}}\}$, Algorithm~\ref{alg:greedy} `greedily' selects the highest weight feasible edge in each iteration---this step is repeated until all available edges that are feasible are added to $G^\ast$. Our main result in this section is that for any given instance of the capacitated matching problem, the matching $G^\ast$ returned by Algorithm~\ref{alg:greedy} has a total weight that is at least
1/3--rd that of the maximum weight matching. 

\begin{theorem}
	For any given capacitated matching problem instance $ \{ \mc{P}, \mc{C}, \bs{b},  (w(a,i))_{(a,i) \in \mc{P}}\}$, let $G^\ast$ denote the output of
Algorithm~\ref{alg:greedy} and $M^\ast$ be any other feasible solution to the
optimization problem in~\eqref{eq:lp_matching} including the optimum matching. Then, $\textstyle\sum_{(a,i) \in M^\ast}w(a,i) \leq 3\sum_{(a,i) \in G^\ast}w(a,i).$
\label{thm_matchapprox}
\end{theorem}
The proof is based on a \emph{charging argument} that takes into account the capacity constraints and can be found in Section~\ref{app:matching}
of the supplementary material. At a high level, we take each edge belonging to the benchmark $M^\ast$ and identify a corresponding edge in $G^\ast$ whose weight is larger than that of the benchmark edge. This allows us to charge the weight of the original edge to an edge in $G^\ast$. During the charging process, we ensure that no more than three edges in $M^\ast$ are charged to each edge in $G^\ast$. This gives us an approximation factor of three.

\subsection{PROPERTIES OF GREEDY MATCHINGS}
We conclude this section by providing a hierarchical decomposition of the edges in $\mc{P}$ for a fixed instance $ \{ \mc{P}, \mc{C}, \bs{b},  (w(a,i))_{(a,i) \in \mc{P}}\}$. In Section~\ref{sec:regretdecomp}, we will use this property to reconcile the offline version of the problem with the online bandit case. Let $G^\ast = \{g^\ast_1, g^\ast_2, \ldots, g^\ast_m\}$ denote the matching
computed by Algorithm~\ref{alg:greedy} for the given instance such that $w(g^\ast_1) \geq w(g^\ast_2) \geq \ldots \geq w(g^\ast_m)$ without loss of
    generality\footnote{If $g=(a,i)$, we abuse notation and let $w(g) =
    w(a,i)$.}. Next, let $G^\ast_j = \{ g^\ast_1, g^\ast_2, \ldots,
    g^\ast_j  \}$ for all $1 \leq j \leq m$---i.e.~the $j$ highest-weight edges in the greedy matching. 

For each $1 \leq j \leq m$, we define the infeasibility set $H^{G^\ast}_j$ as
the set of edges in $\mc{P}$ that when added to $G^\ast_j$ violates the
feasibility constraints of~\eqref{eq:lp_matching}. Finally, we use $L^{G^\ast}_j$ to denote the marginal infeasibility sets---i.e.
$L^{G^\ast}_1 = H^{G^\ast}_1$ and 
\begin{equation}
\label{eqn_infeasibilityset}
L^{G^\ast}_j = H^{G^\ast}_j \setminus H^{G^\ast}_{j-1}, \ \forall \ 2
\leq j \leq m.
\end{equation} 
We note that the marginal infeasibility sets denote a mutually exclusive
partition of the edge set minus the greedy matching---i.e., $\bigcup_{1 \leq j\leq  m}L^{G^\ast}_j = \mc{P} \setminus G^\ast$. Moreover, since the greedy matching selects its edges in the decreasing order of weight, for any $g^\ast_j \in G^\ast$, and every $(a,i) \in
L^{G^\ast}_j$, we have that $w(g^\ast_j) \geq w(a,i)$.

Armed with our decomposition of the edges in $\mc{P} \setminus G^\ast$, we now present a crucial structural lemma. The following lemma identifies sufficient conditions on the \emph{local ordering} of the edge weights for two different instances under which the outputs of the greedy matching for the instances are non-identical. 

\begin{lemma}
Given instances $\{ \mc{P}, \mc{C}, \bs{b},  (w(a,i))_{(a,i) \in \mc{P}}\}$ and $\{ \mc{P}, \mc{C}, \bs{b},  (\tilde{w}(a,i))_{(a,i) \in \mc{P}}\}$   of the capacitated matching problem, let  $G^\ast = \{g^\ast_1, g^\ast_2,
\ldots, g^\ast_m\}$ and $\tilde{G}$ denote the output of
Algorithm~\ref{alg:greedy} for these instances, respectively. Let $E_1, E_2$ be conditions described as follows:
\begin{align*}
E_1 =& \{\exists j < j' \ | (\tilde{w}(g^\ast_j) < \tilde{w}(g^\ast_{j'})) \land
(g^\ast_{j'} \in \tilde{G})\}\\
 E_2 = &  \{\exists g^\ast_j \in G^\ast, (a,i) \in L^{G^\ast}_j\ | \\ &\quad (\tilde{w}(g^\ast_j) < \tilde{w}(a,i)) \land ((a,i) \in \tilde{G})\}.
\end{align*} If $G^\ast \neq \tilde{G}$, then at least one of $E_1$ or $E_2$ must be true.
\label{lem_greedyordering_main}
\end{lemma}
Lemma~\ref{lem_greedyordering_main} is fundamental in the analysis of our MG-EUCB algorithm because it provides a method to map the selection of each sub-optimal edge to a familiar condition comparing empirical rewards to stationary rewards.

\section{ONLINE MATCHING---BANDIT ALGORITHM} 
\label{sec:bandit}
In this section, we propose a multi-armed bandit algorithm for the capacitated
matching problem and analyze its regret.
For concreteness, we first highlight the
information and action sets available to the designer in the online problem. The designer is presented with a \emph{partial instance} of the
matching problem without the weights, i.e.,  
$\{ \mc{P}, \mc{C}, \bs{b}\}$ along with a fixed time horizon of $n$ epochs but
has the ability to set the parameters $(\tau_1, \tau_2, \ldots, \tau_n)$, where
$\tau_k$ is the number of iterations under epoch $k$. The designer's goal is to
design a policy $\alpha$ that selects a matching $\alpha(k)$ in the $k$--th
epoch that is a feasible solution for~\eqref{eq:lp_matching}. At the end of the
$k$--th epoch, the designer observes the average reward $\bs{r}^{\theta_a(k)}_{a,i}$
for each $(a,i) \in \alpha(k)$ but \emph{not the agent's type}. We abuse notation and take $\theta_a(k)$ to be the agent's state at the beginning of epoch $k$. The designer's objective is to minimize the regret over the finite horizon. 

The expected regret of a policy $\alpha$ is
the difference in the expected aggregate reward of a benchmark matching and that of the matching returned by the policy, summed over $n$ epochs. Owing to its favorable properties (see Section~\ref{sec:offline}), we use the greedy matching on the stationary state rewards as our benchmark. Measuring the regret with respect to the unknown stationary-distribution is standard with MDPs (e.g., see \cite{tekin2010online, tekin:2012aa, gai:2011aa}). Formally, let $G^\ast$ denote the output of Algorithm~\ref{alg:greedy} on the instance 
$ \{ \mc{P}, \mc{C}, \bs{b}, (\mu_{a,i})_{(a,i) \in \mathcal{P}} \}$---i.e.,
with the weights $w(a,i)$ equal the stationary state rewards $\mu_{a,i}$. 

\begin{definition}
	The expected regret of a policy $\alpha$ with respect to the greedy matching $G^*$ is given by
	\begin{align*}
\textstyle	R^\alpha(n) & =n\sum_{(a,i) \in G^\ast}\mu_{a,i} -
\sum_{k=1}^n\sum_{(a,i) \in \alpha(k)}\mb{E}[\bs{r}^{\theta_a(k)}_{a,i}], \end{align*}
	where the expectation is with respect to the reward and the state of the agents during each epoch. 
    \label{defn:regret}
\end{definition}

\subsection{REGRET DECOMPOSITION}
As is usual in this type of analysis, we start by decomposing the regret in terms of the number of selections of each sub-optimal arm (edge). 
We state some assumptions and define notation before proving our generic regret
decomposition theorem. A complete list of the notation used can be found in Section~\ref{app:notation} of the supplementary material.
\begin{enumerate}[itemsep=-5pt, topsep=-5pt,leftmargin=15pt]
	\item For analytic convenience, we assume that the number of agents and
        incentives is balanced and therefore, $|\mathcal{A}| = |\mathcal{I}| =
        m$. WLOG, every agent is matched to some incentive in $G^\ast$; if this is not the case, we can add \emph{dummy incentives} with zero reward.
	
	\item Suppose that $G^\ast = \{g^\ast_1, g^\ast_2, \ldots, g^*_m\}$ such that $\mu_{g^*_1} \geq \ldots \geq \mu_{g^*_m}$ and let $i^*(a)$ denote the incentive that $a$ is matched to in $G^*$. Let $L^*_1, \ldots L^*_m$ be the marginal infeasibility sets as defined in~\eqref{eqn_infeasibilityset}. 
	
	\item Suppose that $\tau_0\geq 1$ and $\tauk=\tau_0 + \zeta k$ for some non-negative integer $\zeta$. 
\end{enumerate}

Let $\mathds{1}\{\cdot\}$ be the indicator
 function---e.g., $\mathds{1}\{(a,i) \in \alpha(k)\}$ is one when the edge $(a,i)$ belongs to the matching $\alpha(k)$, and zero otherwise. 
Define $T_{a,i}^\alpha(n)=\sum_{k=1}^{n}\mathds{1}\{(a,i) \in \alpha(k)\}$ to be the random variable that denotes the number of epochs in which an edge is selected under an algorithm $\alpha$. By relating $\mb{E}[T_{a,i}^\alpha(n)]$
to the regret $R^\alpha(n)$, we are able to provide bounds on the performance of $\alpha$. 

By  adding and subtracting $\sum_{(a,i)\in
        \mc{P}}\mb{E}[T^\alpha_{a,i}(n)]\mu_{a,i}$ from the equation in
        Definition~\ref{defn:regret}, we get that 
\begin{align*}
   & R^\alpha(n) = \textstyle\sum_{(a,i)\in \mc{P}}\mb{E}[T_{a,i}^\alpha(n)](\mu_{a,i^*(a)}-\mu_{a,i})\nonumber \\ & \textstyle+\sum_{k=1}^n\sum_{(a,i)\in
        \mc{P}}\mb{E}[\mathds{1}\{(a,i) \in \alpha(k)\}\big(
        \mu_{a,i}-\bs{r}^{\theta_a(k)}_{a,i} \big)].
\end{align*}
To further simplify the regret, we separate the edges in $\mathcal{P}$ by
introducing the notion of a sub-optimal edge. Formally, for any given $a \in
\mc{A}$, define $S_a \coloneqq \{ (a,i) ~|~ \mu_{a,i^*(a)} \geq \mu_{a,i}
~\forall i \in  \mc{I}\}$ and $\mc{S} := \bigcup_{a \in \mc{A}} S_a$. Then, the regret bound in the above equation can be simplified by ignoring the contribution of the terms in $\mc{P} \setminus \mc{S}$. That is, since $\mu_{a,i^*(a)} < \mu_{a,i}$ for all $(a,i) \in \mc{P} \setminus \mc{S}$,
\begin{align}
 &  R^\alpha(n)   \leq  \textstyle\sum_{(a,i)\in
       \mc{S}}\mb{E}[T_{a,i}^\alpha(n)](\mu_{a,i^*(a)}-\mu_{a,i}) \notag \\
    &\textstyle+\sum_{k=1}^n\sum_{(a,i)\in
    	\mc{P}}\mb{E}[\mathds{1}\{(a,i) \in \alpha(k)\}\big(
    \mu_{a,i}-\bs{r}^{\theta_a(k)}_{a,i} \big)].    \label{eqn_regret_2}
\end{align}
Recall from the definition of the marginal infeasibility sets in~\eqref{eqn_infeasibilityset} that for any given $(a,i) \in \mc{P} \setminus G^*$, there exists a unique edge $g^*_j \in G^*$ such that $(a,i) \in L^*_j$. Define $L^{-1}(a,i) \coloneqq g^*_j \in G^*$ such that $(a,i) \in L^*_j$. Now, we can define the reward gap for any given edge as follows: 
\begin{equation*}
 \Delta_{a,i}=  \left\{ \begin{array}{ll}
 \mu_{a,i^*(a)}- \mu_{a,i}, & \text{if $(a,i) \in \mc{S}$}\\
\mu_{L^{-1}(a,i)} - \mu_{a,i}, & \text{if $(a,i) \in (\mc{P} \setminus G^*) \setminus \mc{S}$}\\
\mu_{g^*_{j-1}} - \mu_{g^*_j}, &\text{if $(a,i) = g^*_j$ for $j \geq 2$}
\end{array}\right.
\end{equation*}

This leads us to our main regret decomposition result which leverages mixing times for Markov chains \cite{fill:1991aa} along with Equation~\eqref{eqn_regret_2} in deriving regret bounds. For an aperiodic,
	irreducible Markov chain $P_{a,i}$, 
  using the notion that it convergences to its stationary state under repeated plays of a fixed action, we can prove that for every arm $(a,i)$, there exists a constant $C_{a,i}>0$ such that $\big|\mb{E}\big[ \mu_{a,i}-\bs{r}^{\theta_a(k)}_{a,i}\big]\big|\leq
    C_{a,i}/\tauk$---in fact, this result holds for all type distributions $\beta_a^{(k)}$ of the agent.
\begin{proposition} Suppose for each $(a,i)\in \mc{P}$, $P_{a,i}$ is an aperiodic,
	irreducible Markov chain with corresponding constant $C_{a,i}$.
	Then, for a given algorithm $\alpha$ where $\tau_k=\tau_0+\zeta k$ for some
	fixed $\zeta>0$, we have that
	\begin{align*}
	 R^\alpha(n) \leq & \textstyle\sum_{(a,i) \in \mc{S} }\mb{E}_\alpha\big[ T_{a,i}^\alpha(n)
	\big](\Delta_{a,i}  +\frac{C_{a,i}}{\tau_0}) \\
    &\textstyle+m\frac{C_{_\ast}}{\zeta}\big(
	1+\log\big( \zeta(n-1)/\tau_0+1 \big)\big).
	\end{align*}
	\label{prop:regretdecomp1}
\end{proposition}
The proof of this proposition is in Section~\ref{app:proof:regretdecomp} of the supplementary material.

\label{sec:regretdecomp}

\begin{figure*}[t]
    \centering
    \subfloat[]{\includegraphics[width=.25\textwidth]{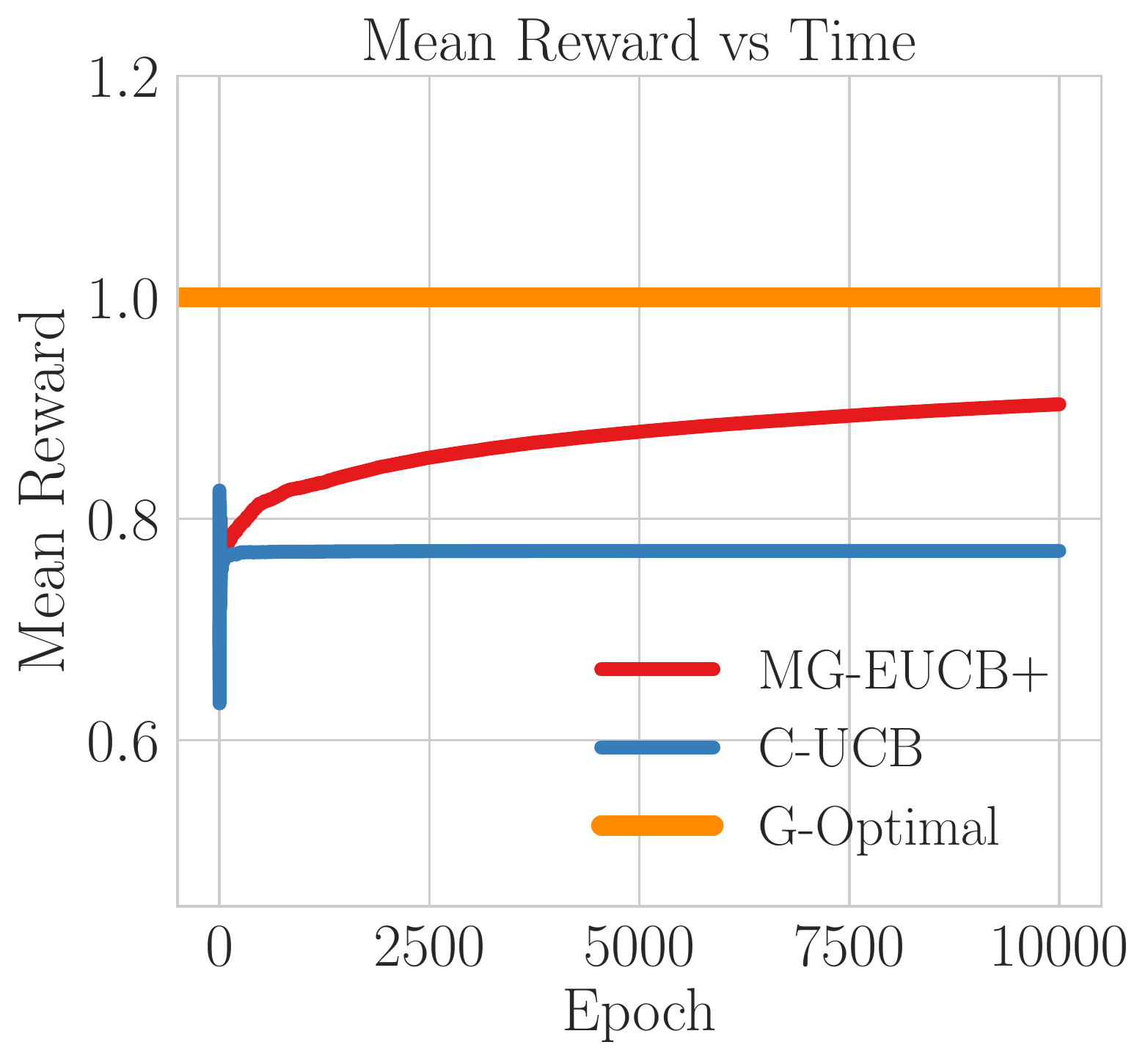}\label{fig:epoch_vs_classic}}
    \hfill\subfloat[]{\includegraphics[width=.25\textwidth]{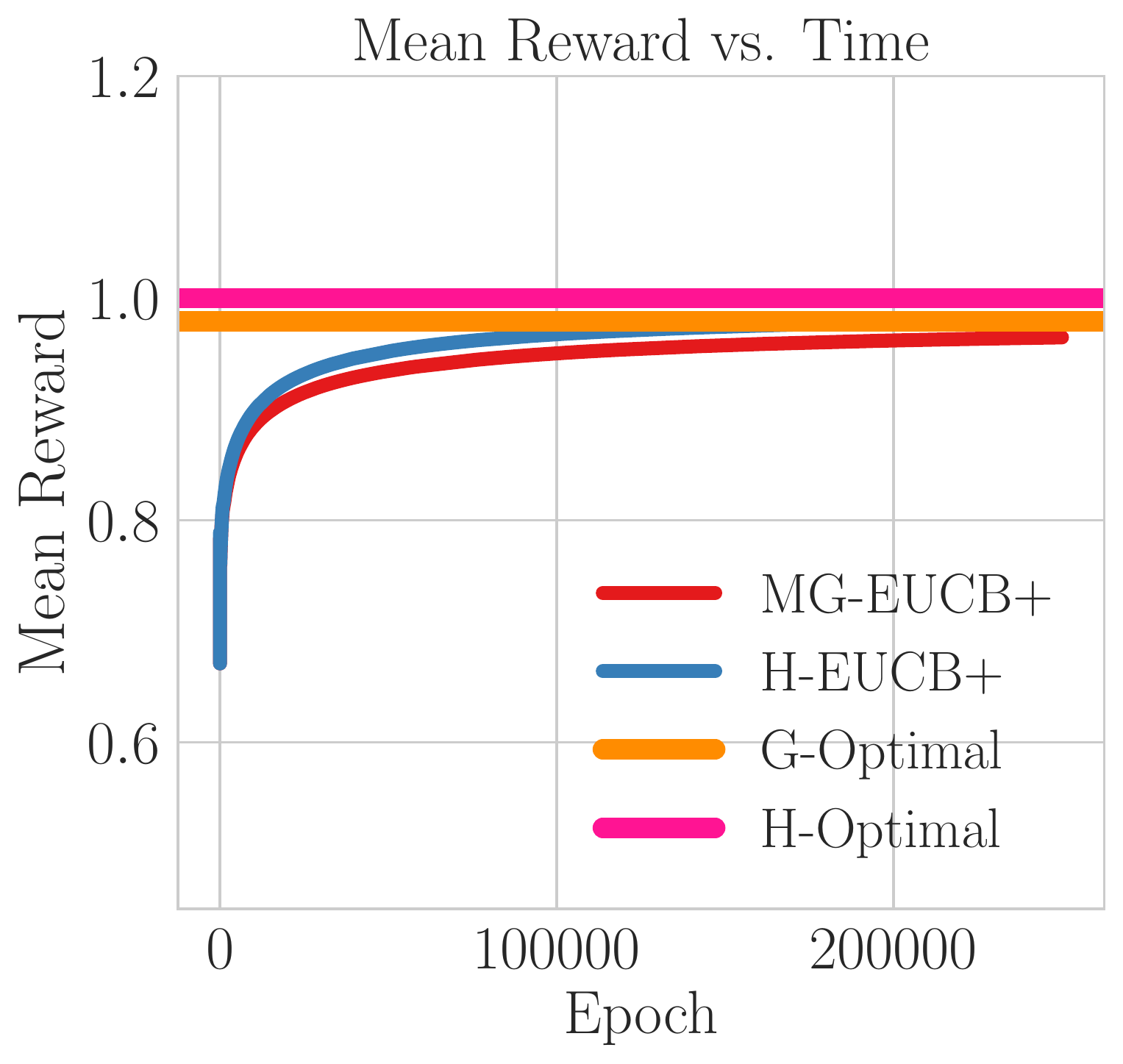}\label{fig:epoch_mean_reward}}
    \hfill\subfloat[]{\includegraphics[width=.25\textwidth]{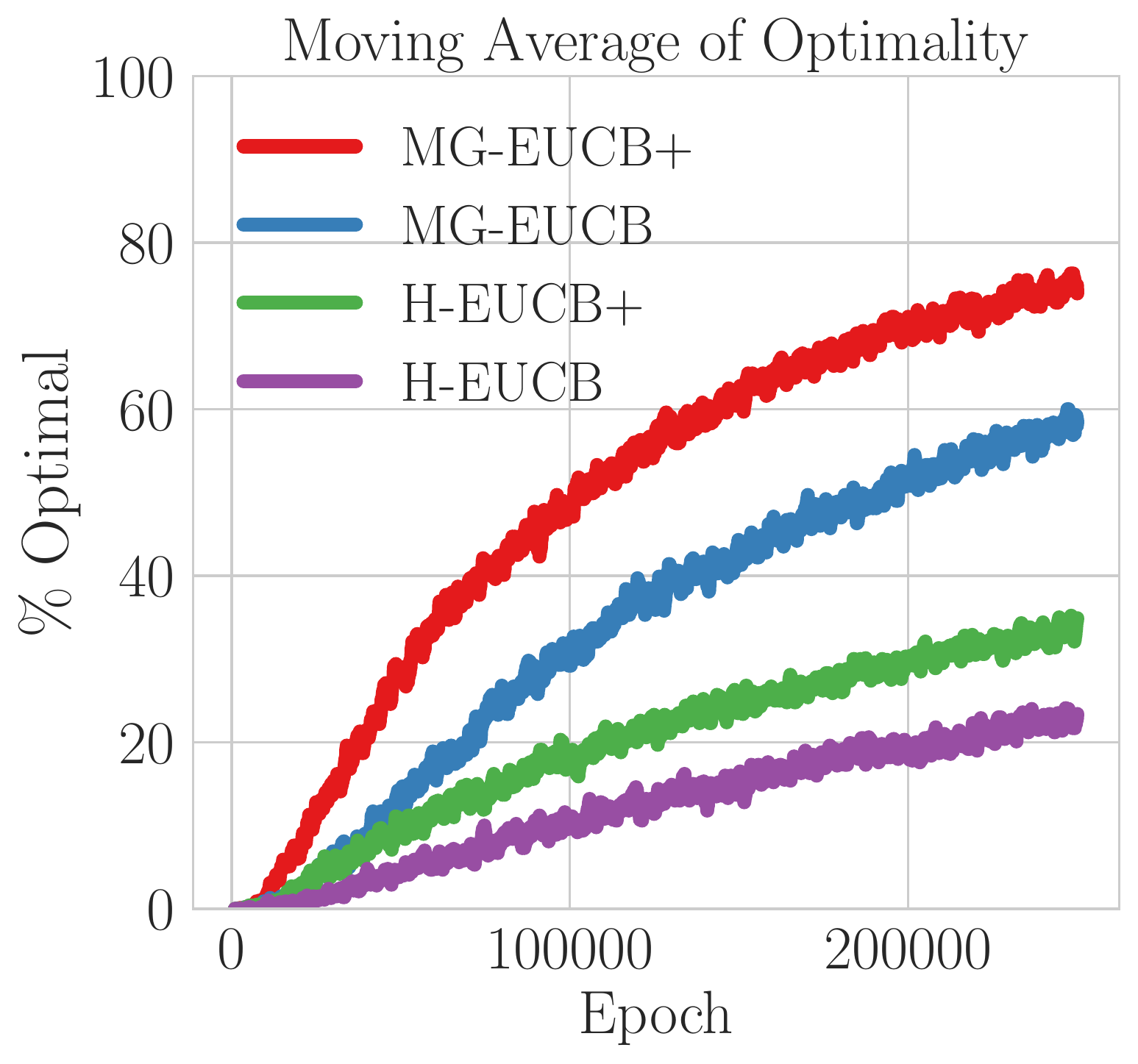}\label{fig:ma_optimal}}
     \caption{Synthetic Experiments: Comparison of
     MG-EUCB(+)
 and H-EUCB(+) to their respective offline solutions (G- and H-optimal,
 respectively) and to C-UCB (classical UCB). We use the following set up: (i) ${|\mathcal{A}| = |\mathcal{I}| =
|\Theta_a| = 10}$ (see Supplement~\ref{app:additional} for more extensive
experiments) (ii) each state transition matrix $P_{a,i}$
associated with an arm $(a, i) \in \mathcal{P}$ was selected uniformly at random within the
class of aperiodic and irreducible stochastic matrices; (iii) the reward for each arm, state pair $r^\theta_{a, i}$ is drawn i.i.d.~from a
distribution $\mc{T}_r(a, i, \theta)$ belonging to either a Bernoulli, Uniform,
or Beta distribution; (iv) $\tau_0 = 50$ and $\zeta = 1$.}
\end{figure*}

\subsection{MG-EUCB ALGORITHM AND ANALYSIS}
\label{sec:ucb}
In the initialization phase, the algorithm computes and plays a sequence of
matchings $M_1, M_2, \ldots, M_\p$ for a total of $p$ epochs. The initial
matchings ensure that every edge in $\mc{P}$ is selected at least once---the
computation of these initial matchings relies on a \emph{greedy covering}
algorithm that is described in Section~\ref{app:permutematchings} of the supplementary material. Following
this, our algorithm maintains the cumulative empirical reward $\bar{r}_{a,i}$ for every
$(a,i) \in \mc{P}$. At the beginning of (say) epoch $k$, the
algorithm computes a greedy matching for the instance  $\{ \mc{P}, \mc{C},
\bs{b},  (w(a,i))_{(a,i) \in \mc{P}}\}$ where $w(a,i) = \bar{r}_{a,i}/T_{a,i} +
c_{a,i}$, i.e., the average empirical reward for the edge added to a suitably
chosen confidence window. The \textproc{incent}$(\cdot)$ function
(Algorithm~\ref{alg:armpull}, described in the supplementary material since it is a trivial function) plays each edge in the greedy matching for
$\tau_k$ iterations, where $\tau_k$ increases linearly with $k$. This process is repeated for $n - \p$ epochs. Prior to theoretically analyzing MG-EUCB, we return to Example~\ref{ex:cuteex} in order to provide intuition for how the algorithm overcomes correlated convergence of rewards.
\begin{algorithm}[t]
    \caption{MatchGreedy-EpochUCB}
    \label{alg:ucb}
    \begin{algorithmic}[1] 
        \Procedure{MG-EUCB}{$\zeta$, $\tau_0, \mc{P}$}\\ 
        \spaceit $t_1\gets 0$, $\bar{r}_{a,i}\gets 0$ \& $T_{a,i}\gets 1$ $\
        \forall (a,i) \in \mc{P}$\\
        $\ \ $$M_1, M_2, \ldots, M_{\p}\subset\mc{P}$ s.t.~$(a,i)
        \in M_j \Leftrightarrow (a,i)\notin M_\ell\ \forall \ell\neq j$ 
        \Comment{see
            Supplement~\ref{app:permutematchings} for details}\\
            \spaceit \Call{incent}{$\cdot$} \Comment{see
                Alg.~\ref{alg:armpull} in Supplement~\ref{app:algs}}\\
        \spaceit\textbf{for}{$1\leq n\leq m$} \Comment{play each arm once}\\
             $\quad\ $$(\bar{r}_{a,i})_{(a,i) \in M_n}\gets$ \Call{incent}{$M_n$, $t_n$, $n$, $\tau_0$, $\zeta$} \\
            $\quad\ $$t_{n+1}\gets t_n+\tau_0+\zeta n$\\ 
                \spaceit\textbf{end for}\\
                \spaceit\textbf{while}{ $n>m$}\\
                $\quad\ $
                $M_G=\Call{MG}{(\bar{r}_{a,i}/T_{a,i}+c^{T_{a,i}}_{a,i}(n))_{(a,i)
                    \in \mc{P}}$} \\
            $\quad\ $$(r_{a,i}(t_n))_{(a,i) \in M_G}\gets$
            \Call{incent}{$M_G,t_n,n,\tau_0,\zeta$} \\
            $\quad\ $$\bar{r}_{a,i}\gets \bar{r}_{a,i}+\frac{1}{\tau_0+\zeta n}r_{a,i}(t_n)$ ~$\forall (a,i) \in M_G$
            \\
            $\quad\ $$T_{a,i}\gets T_{a,i}+1~\forall (a,i) \in M_G$\\
            $\quad \ $$t_{n+1}\gets t_n+\tau_0+\zeta n$; $n\gets
            n+1$\\
            \spaceit\textbf{end while}\label{euclidendwhile}
        \EndProcedure
    \end{algorithmic}
\end{algorithm}

\textbf{Revisiting Example 1:}  Why does MG-EUCB work?  In Example 1, the algorithm initially estimates the empirical reward of $(a_1, i_i)$ and $(a_2, i_2)$ to be zero respectively. However, during the UCB exploration phase, the matching $M_1 = {(a_1, i_1), (a_2, i_2)}$ is played again for epoch length $> 1$ and the state of agent $a_1$ moves from $\theta_1$ to $\theta_2$ during the epoch. Therefore, the algorithm estimates the average reward of each edge within the epoch to be $\geq 0.5$, and the empirical reward increases. 
This continues as the epoch length increases, so that eventually the empirical reward for $(a_1, i_1)$ exceeds that of $(a_1, i_2)$ and the algorithm correctly identifies the optimal matching as we move from exploration to exploitation.  
	
In order to characterize the regret of the MG-EUCB algorithm, Proposition~\ref{prop:regretdecomp1} implies that it is sufficient to bound the expected number of epochs in which our algorithm selects each sub-optimal edge. The following theorem presents an upper bound on this quantity. 

\begin{theorem}
Consider a finite set of $m$ agents $\mc{A}$ and incentives $\mc{I}$ with corresponding aperiodic,
irreducible Markov chains $P_{a,i}$ for each $(a,i)\in \mc{P}$.   Let $\alpha$ be the MG-EUCB algorithm with mixing time sequence
$\{\tau_k\}$ where $\tau_k=\tau_0+\zeta k$, $\tau_0>0$, and $\zeta>0$.
  Then for every $(a,i) \in \mc{S}$, 
   \begin{align*}
   \mb{E}_\alpha[T_{a,i}(n)]&\leq \textstyle\frac{4m^2}{\Delta_{a^*,i^*}^2}\left(
       \frac{\rho_{a^*,i^*}}{\sqrt{\tau_0}}+\sqrt{6\log n + 4\log m}
       \right)^2\notag\\
       &+2(1+\log(n))
   \end{align*}
where $(a^\ast, i^\ast) = \argmax\limits_{(a_1, i_1) \in \mc{P} \setminus
g^\ast_1} \Big\lceil \frac{4}{\Delta_{a_1,i_1}^2}\big(
\frac{\rho_{a_1,i_1}}{\sqrt{\tau_0}} +\sqrt{6\log n + 4\log m} \big)^2\Big\rceil$, and $\rho_{a,i}$ is a constant specific to edge $(a,i)$.
    \label{thm:regretbound}
\end{theorem}

The full proof of the theorem is provided can be found in the supplementary material.
\begin{proof}(sketch.)
There are three key ingredients to the proof:  (i) linearly increasing epoch
lengths, (ii) overcoming cascading errors, and (iii) application of the
Azuma-Hoeffding concentration inequality. 

By increasing the epoch length linearly, MG-EUCB ensures that as the algorithm
converges to the optimal matching, it also plays each arm for a longer duration
within an epoch. This helps the algorithm to progressively discard sub-optimal
arms without selecting them too many times when the epoch length is still small. At the same time, the epoch length is long enough to allow for sufficient mixing and separation between multiple
near-optimal matchings. 
If we fix the epoch length as a constant, the resulting regret bounds are considerably worse because the agent states may never converge to the steady-state distributions.

To address cascading errors, we provide a useful characterization.
For a given $(a,i)$,
 suppose that $u^k_{a,i}(t)$ refers to the average
    empirical reward obtained from edge $(a,i)$ up to epoch $t-1$ plus the
    upper confidence bound parameter, given that edge $(a,i)$ has been selected
    for exactly $k$ times in epochs $1$ to $t-1$  
    . 
	For any given epoch $k$ where the algorithm selects a sub-optimal matching, i.e., $\alpha(k) \neq G^\ast$, we can apply Lemma~\ref{lem_greedyordering_main} and get that  at least one of the following conditions must be true:
    \begin{enumerate}[itemsep=-5pt,topsep=-2pt, leftmargin=15pt]
\item $\mathds{1}\{\exists j < j'|\ \big(u_{g^\ast_{j'}}^k(t) >
    u_{g^\ast_j}^k(t)\big) \land ( g^\ast_{j'} \in \alpha(t))\}$
\item $\mathds{1}\{\exists j, (a,i) \in L^\ast_j|\ \big( u_{g^*_{j}}^k(t) <
    u_{a,i}^k(t)\big) \vee ((a,i) \in \alpha(k))    \} =1$
	\end{enumerate}
\begin{figure*}[t]
    \centering
    \subfloat[Static Demand]{\includegraphics[width=.25\textwidth]{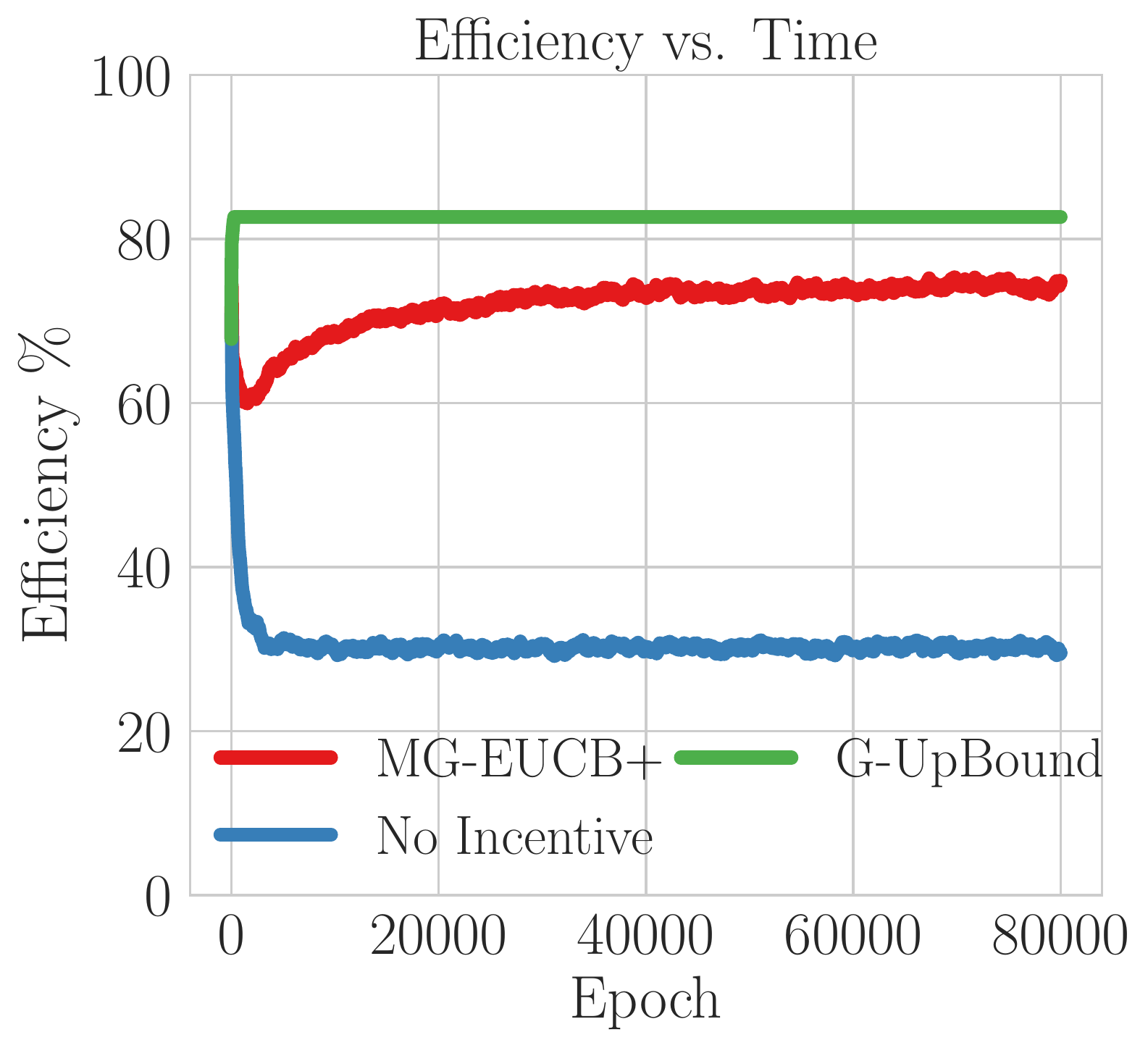}\label{fig:eff_static}}\hfill
    \subfloat[Random Demand]{\includegraphics[width=.25\textwidth]{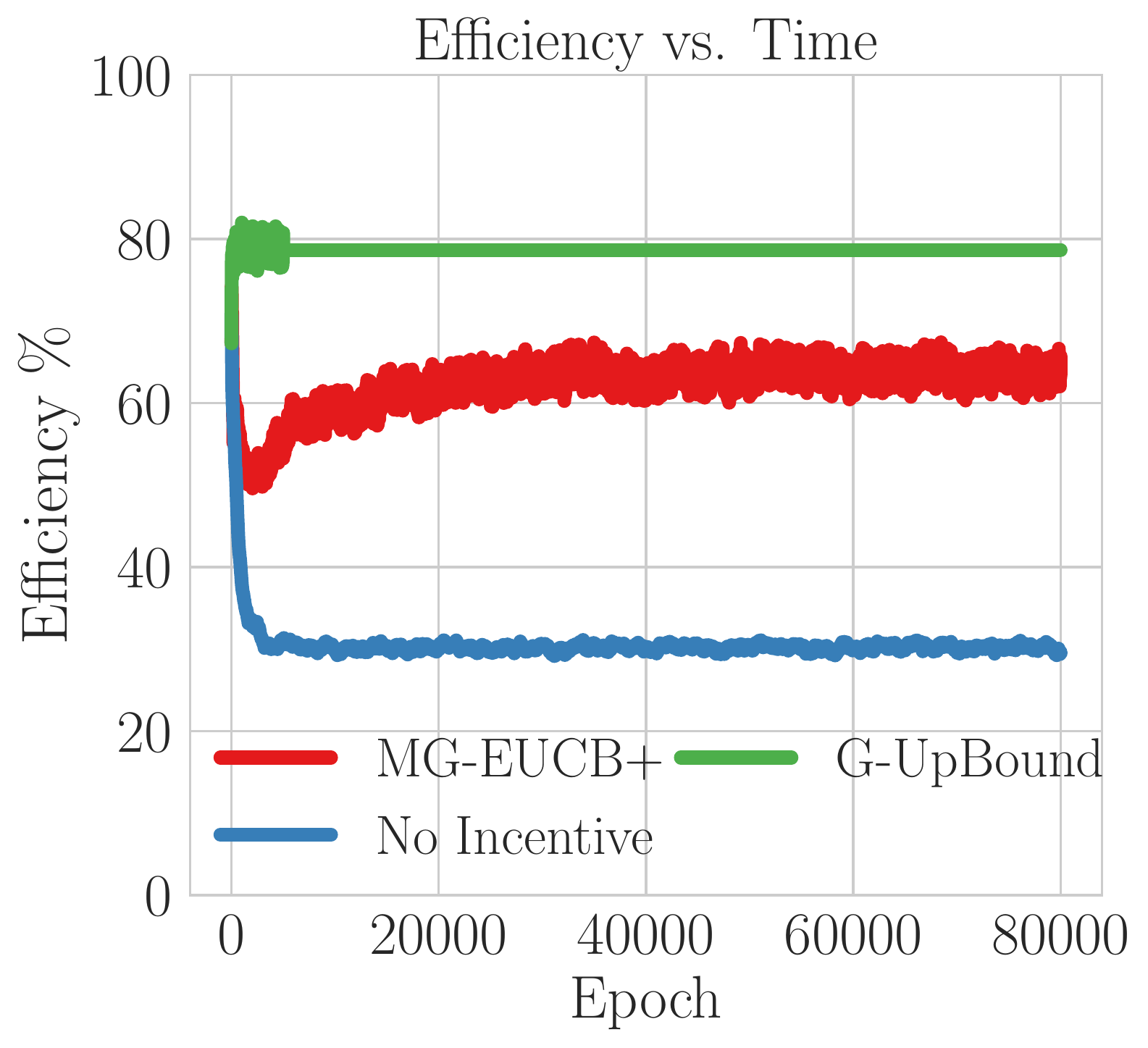}\label{fig:eff_random}}\hfill
    \subfloat[]{\includegraphics[width=.25\textwidth]{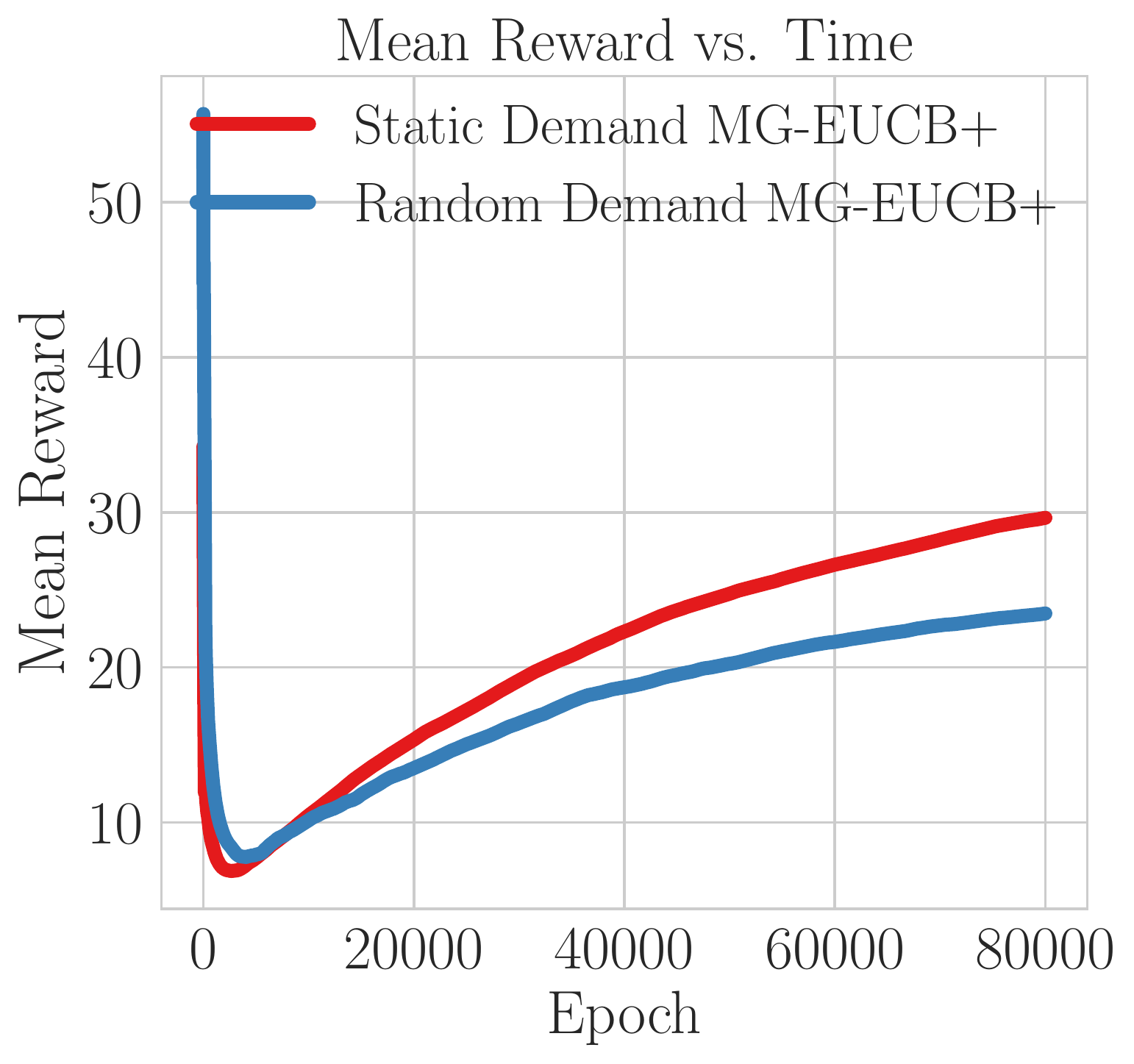}\label{fig:mean_both}}
     \caption{Bike-share Experiments: Figures~\ref{fig:eff_static} and~\ref{fig:eff_random} compare the efficiency (percentage of demand satisfied) of the bike-share system with two demand models under incentive matchings selected by MG-EUCB+ with upper and lower bounds given by the system performance when the incentives are computed via the benchmark greedy matching that uses the state information and when no incentives are offered respectively. In Figure~\ref{fig:mean_both} we plot the mean reward of the MG-EUCB+ algorithm with static and random demand which gives the expected number of agents who accept an incentive within each epoch.}
    \label{fig:bike_figs}
\end{figure*}

    This is a particularly useful characterization because it maps the selection of each sub-optimal edge to a familiar condition that compares the empirical rewards to the stationary rewards. Therefore, once each arm is selected for $O(\log(n))$ epochs, the empirical rewards approach the `true' rewards and our algorithm discards sub-optimal edges. Mathematically, this can be written as
   \begin{align*}
       &\mb{E}_\alpha[T_{a',i'}(n)] =\textstyle 1+\sum_{t=p+1}^n\mathds{1}\{(a',i') \in \alpha(t)\}\\
    & \leq    \textstyle \ell m^2 +\sum_{j=1}^m
    \sum_{(a,i) \in L^+_j}   \textstyle \sum_{t=p+1}^n\sum_{s=1}^{t-1}\sum_{k=\ell}^{t-1}\big(\\&    \textstyle  
   \textstyle 
    \mathds{1}\{u^s_{g^*_j}(t) \leq  u^k_{a,i}(t)  \}\big),
    \end{align*}
       where $\ell$ is some carefully chosen constant, $L^+_j = L^*_j \cup \{g^\ast_{j+1} \}$ and $L^+_m = L^\ast_m$. 
       
       With this
       characterization, for each $s$, we find an upper bound on the probability of the event
       $\{u^s_{g^*_j}(t) \leq  u^k_{a,i}(t)\}$. However, this is a non-trivial task since the reward obtained in any given epoch is \emph{not independent} of the previous actions. Specifically, the underlying Markov process that generates the rewards is common across the edges connected to any given agent in the
       sense, that the initial distribution for each Markov chain that results
       from pulling an edge is the distribution at the end of the preceding
       pull. Therefore, we employ
       Azuma-Hoeffding~\cite{azuma:1967aa,hoeffding:1963aa},  a concentration inequality that does not require independence in the arm-based observed rewards. 
              Moreover, unlike the classical UCB analysis, the empirical reward can differ
from the expected stationary reward due to the
distributions $\mc{T}_r(a,i,\theta)$ and 
$\beta^k_{a,i} \neq \pi_{a,i}$. To account for this additional error term, we 
use bounds on the convergence rates of Markov chains to guide the choice of the confidence parameter $c^k_{a,i}(t)$ in
Algorithm~\ref{alg:ucb}. Applying the Azuma-Hoeffding inequality, we can show that with high probability, the difference between the empirical reward and the stationary reward of edge $(a,i)$ is no larger than $c^k_{a,i}(t)$. \end{proof} 

As a direct consequence of Proposition~\ref{prop:regretdecomp1} and
Theorem~\ref{thm:regretbound}, we get that for a fixed instance, the regret only
increases logarithmically with $n$. 



\section{EXPERIMENTS}
\label{sec:experiments}

In this section, we present a set of illustrative experiments with our algorithm (MG-EUCB) on synthetic and real data. We observe much faster convergence
with the greedy matching as compared to the Hungarian algorithm. Moreover, as is typical in the bandit literature (e.g.,~\cite{auer:2002aa}), we show that a \textit{tuned} version of our algorithm (MG-EUCB+), in which we reduce the coefficient on the $\log(n)$ term
in the UCB `confidence parameter' from six to three, further improves the convergence of our algorithm. Finally we show that our algorithm can be effectively used as an incentive design scheme to improve the performance of a bike-share system.

\subsection{SYNTHETIC EXPERIMENTS}
We first highlight the failure of classical UCB approaches (C-UCB)---e.g., as in~\cite{gai:2011aa}---for problems with correlated reward evolution. In Figure~\ref{fig:epoch_vs_classic}, we demonstrate that C-UCB converges
almost immediately to a suboptimal solution, while this is not the case for our algorithm (MG-EUCB+). In Figure~\ref{fig:epoch_mean_reward}, we compare MG-EUCB and MG-EUCB+ with a variant of Algorithm~\ref{alg:ucb} that uses the Hungarian method (H-EUCB) for matchings. While H-EUCB does have a `marginally' higher mean reward,
Figure~\ref{fig:ma_optimal} reveals that the MG-EUCB and MG-EUCB+ algorithms converge much faster to the optimum solution
of the greedy matching than the Hungarian alternatives.

\subsection{BIKE-SHARE EXPERIMENTS}
In this problem, we seek to incentivize participants in a bike-sharing system;
our goal is to alter their intended destination in order to balance the spatial
supply of available bikes appropriately and meet future user demand. We use data
from the Boston-based bike-sharing service Hubway \cite{hubway} to construct the
example. Formally, we consider matching each agent $a$ to an incentive $i = s_a'$, meaning the algorithm proposes that agent $a$ travel to station
$s'_a$ as opposed to its intended destination $s_a$ (potentially, for some monetary benefit). The agent's state $\theta_a$ controls the probability of accepting the incentive by means of a distance threshold parameter and a parameter of a Bernouilli distribution, both of which are drawn uniformly at random. More details on the data and problem setup can be found in Section~\ref{app:additional} of the supplementary material.

Our bike-share simulations presented in Figure~\ref{fig:bike_figs} show approximately a $40$\% improvement in system performance when compared to an environment without incentives and convergence towards an upper bound on system performance. Moreover, our algorithm achieves this significant performance increase while on average matching less than $1$\% of users in the system to an incentive. 

\section{Conclusion}
\label{sec:conclusion}
We combine ideas from greedy matching, the UCB multi-armed bandit strategy, and the theory of Markov chain mixing times to propose a bandit algorithm for matching incentives to users, whose preferences are unknown a priori and evolving dynamically in time, in a resource constrained environment. For this algorithm, we derive logarithmic gap-dependent regret bounds despite the additional technical challenges of cascading sub-optimality and correlated convergence. Finally, we demonstrate the empirical performance via examples.

\subsubsection*{Acknowledgments}
\label{sec:acks}
This work is supported by NSF Awards CNS-1736582 and CNS-1656689. T. Fiez was also supported in part by an NDSEG Fellowship.

\clearpage
\bibliographystyle{plainnat}
\bibliography{2018uai}

\clearpage

\appendix
\ifdsfonts
\section{NOTATIONAL TABLE}
\label{app:notation}
	\begin{tabular}{|l|l|}
		\hline
        \textbf{notation} & \textbf{meaning}\\\hline\hline
		$\mc{A}$ & set of agents \\\hline
		$\mathcal{I}$ & set of incentives\\\hline
		$\mc{P}=\mc{A} \times \mc{I}$ & allowed agent-incentive pairs\\\hline
		$\Theta_a$ & state (type) space of agent $a$\\\hline
		$P_{a,i}$ & transition probability kernel \\\hline
    $\beta^{(t)}_a$ & agent $a$'s type distribution at epoch $t$\\\hline
		$\pi_{a,i}$ & stationary distribution of $(a,i) \in \mc{P}$\\\hline
        \multirow{1}{*}{$\mu_{a,i}$} & expected reward from $(a,i) \in
        \mc{P}$\\\hline
         \multirow{2}{*}{$\tau_k$} & number of iterations matching offered\\
         & in epoch $k$, $\tau_k=\tau_0 + \zeta k$, $\zeta>0$\\\hline
         \multirow{1}{*}{$r_{a,i}^{\theta_a}$} & random  reward \\\hline
         \multirow{1}{*}{$\mc{T}_r(a,i,\theta_a)$} & agent $a$'s reward
         distribution \\\hline
         \multirow{2}{*}{$\bs{r}^{\theta}_{a,i}$} & time-averaged reward during
         epoch $k$\\& $\bs{r}^{\theta_a(k)}_{a,i} =
         \frac{1}{\taua{k}}\sum_{t=t_k}^{t_{k+1}-1}r_{a,i}^{\theta_a(t)}$\\\hline
         \multirow{1}{*}{$b_{\class_l}$} & maximum number of edges of class
         $\xi_l$\\\hline
         {$G^*$} & greedy matching on weights $(\mu_{a,i})$\\\hline
         \multirow{2}{*}{$g^*_j$} & the edge having the $j$--th\\
         &largest weight $(\mu_{a,i})$ in $G^*$.\\\hline
        \multirow{1}{*}{$i^*(a)$} & incentive agent $a$ is matched to in $G^*$\\
        \hline
        \multirow{2}{*}{$L^*_{j}$} & set of $(a,i) \in \mathcal{P}$ that become
        infeasi-\\
        & ble when $g^*_j$ is added to matching\\ &but not before that \\\hline
        \multirow{2}{*}{$S_a$} & set of edges $(a,i)$ such that\\
        &$\mu_{a,i} \leq
        \mu_{a,i^*(a)}$\\\hline
		$\mc{S}$ & $\bigcup_{a \in \mc{A}}S_a$\\\hline
		$m$ & number of agents \& incentives\\\hline
		$n$ & the total number of epochs  \\\hline
        \multirow{2}{*}{$\theta_a(t)$} & state of agent $a$ at the beginning\\
        &of epoch $t$\\\hline
		$C_{a,i}, \rho_{a,i}$ & constants specific to each edge $(a,i)$ \\\hline
		$C_{\ast}$ & $\max_{(a,i) \in \mc{P} \setminus \mc{S}}C_{a,i}$\\\hline
        \multirow{2}{*}{$R^{\alpha}(n)$} & regret of given matching policy\\
        &$\alpha$ at the end of $n$ epochs\\\hline
        \multirow{2}{*}{$T_{a,i}(n)$} & number of times edge $(a,i)$ 
        \\
        & selected in first $n$ epochs\\\hline
        \multirow{2}{*}{$R^{\theta,k}_{a,j}$} & reward on edge $(a,i)$ when
        selected  \\
        &for the $k$--th time given $\theta_a$\\\hline
        \multirow{2}{*}{$\bar{R}^k_{a,j}$} & average reward on first $k$
        times $(a,i)$\\
        &is selected, i.e.,
	     $\frac{1}{k}\sum_{i=1}^{k}R^{\theta,k}_{a,j}$\\\hline
	    \multirow{2}{*}{ $\theta_a(t^l_{a,i})$}& agent $a$'s state at the
        beginning\\
        &of epoch
         $l$\\	\hline
         \multirow{1}{*}{$X_{a,i}^k$} & $R^{\theta,k}_{a,i}-\mb{E}[R^{\theta,k}_{a,i}|\mc{F}^{k-1}_{a,i}]$\\
		$Y_{a,i}^k$ & $\sum_{j=1}^kX_{a,i}^j$ a martingale\\\hline
        \multirow{1}{*}{$Q_{a,i}(k)$} & $\frac{C_{a,i}}{2}\left(\frac{1}{\zeta+\tau_0}+\frac{1}{\zeta}\log\left(
		 1+\frac{k\zeta}{\tau_0} \right) \right)$\\\hline
         \multirow{2}{*}{$c^k_{a,j}(t)$} & upper confidence parameter for edge
         \\
         &$(a,j)$ after being selected for $k$ times \\  \hline
         \multirow{2}{*}{$u^k_{a,j}(t)$} & average reward plus upper
         confidence\\
         &parameter for $(a,j)$, i.e., $\bar{R}^k_{a,j} + c^k_{a,j}$ \\
		\hline
	\end{tabular}

\section{PROOFS}

\subsection{PROOF OF THEOREM 1}
\label{app:matching}
\begin{proof}
Our proof relies on what is referred to in the matching literature as a
\emph{charging argument}. In simple terms, we take each edge belonging to the
benchmark $M^\ast$ and identify a corresponding edge in $G^\ast$ whose weight is
larger than that of the benchmark edge. This allows us to charge the weight of
the original edge to an edge in $G^\ast$. During the charging process, we ensure
that no more than three edges in $M^\ast$ are charged to each edge in $G^\ast$.
This gives us an approximation factor of three.	
	
Suppose that an edge $(a,i)$ belongs to $M^\ast$ but not $G^\ast$. This implies
that the edge $(a,i)$ was removed from the set $E'$ at some iteration during the
course of Algorithm~\ref{alg:greedy}. Moreover, as per the algorithm, this
removal can happen in one of two ways: (i)  via
Line~\ref{alggreedy_firstremoval}, in which case there exists some edge $(a,i')$ or $(a', i)$
that was selected to $G^\ast$ ahead of $(a,i)$, and (ii) via
Line~\ref{alggreedy_secondremoval} in which case $b_{\xi_j}$ edges belonging to
class $\xi_j = c(a,i)$ were added to $G^\ast$ before $(a,i)$, as a result of which
the capacity constraint for that class was met. Based on this, we divide the
analysis into two cases.

\textbf{Case I: Removal via Line~\ref{alggreedy_firstremoval}}.
Without loss of generality, suppose that $(a', i)$ is the edge added to $G^\ast$
during the iteration in which $(a,i)$ is removed. Then, by definition, since
$(a',i) = \arg\max_{(a'', i'') \in E'} w(a'',i'')$ before the removal of $(a,i)$
from $E'$, we infer that 
\begin{equation}
\label{eqn_firstrem}
w(a, i) \leq w(a',i) 
\end{equation} 

\textbf{Case II: Removal via Line~\ref{alggreedy_secondremoval}}.
In this case, since the class $\xi_j = c(a,i)$ has reached its capacity limit, and
since the greedy algorithm selects edges in the decreasing order of weight, it
must be the case that for every $(a',i') \in G^\ast \cap \xi_j$, we have that
\[w(a,i) \leq w(a',i').\]
Since $G^\ast \cap \xi_j$ contains exactly $b_{\xi_j}$, we can average the above
equation over the edges in $G^\ast \cap \xi_j$ to get that 
\begin{equation}
\label{eqn_secondrem}
w(a,i) \leq\textstyle \frac{1}{b_{\xi_j}} \sum_{(a',i') \in G^* \cap \xi_j}w(a',i').
\end{equation}

Finally, we note that if edge $(a,i)$  belongs to both the greedy matching and
$M^\ast$, we can simply `charge the weight of $(a,i)$' to itself. 

Now we can complete the proof by summing \eqref{eqn_firstrem} and
\eqref{eqn_secondrem} over all the edges in $M^\ast$. Formally, let $M^\ast
= M^\ast_1 \cup M^\ast_2$ such that $M^\ast_1$ denotes the set of edges that are
present in both $M^\ast$ and $G^\ast$ as well as the edges that fall under the
first case. Similarly, let $M^\ast_2$ denote the edges that fall under the
second case. Summing \ref{eqn_firstrem} over all of the edges in $M^\ast_1$, we get that
\begin{equation}
\label{eqn_finalgreedy1}
\textstyle\sum_{(a,i) \in M^\ast_1}w(a,i) \leq 2\sum_{(a,i) \in G^\ast}w(a,i).
\end{equation}
The factor of two in the right hand side comes from the fact that for any given
edge $(a,i)$ in $G^\ast$, at most two edges in $M^\ast_1$ can be charged to this
edge. Indeed, the only edges that can be charged to $(a,i)$ must contain either
the node $a$ or the node $i$ and in a matching, each node can appear in at most
one edge. Next, summing~\eqref{eqn_secondrem} over all of the edges in
$M^\ast_2$, we get that
\begin{align}
\textstyle\sum_{(a,i) \in M^\ast_2}w(a,i) &= \textstyle\sum_{\xi_j \in \mc{C}} \sum_{(a,i)
    \in  M^\ast_2
\cap \xi_j}w(a,i)\notag\\
&\leq\textstyle \sum_{\xi_j \in \mc{C}} \sum_{(a,i) \in \xi_j \cap G^\ast
}w(a,i)\notag\\
&= \textstyle\sum_{(a,i) \in G^\ast}w(a,i).
\label{eqn_finalgreedy2}
\end{align}

To see why this is the case, first observe that in~\eqref{eqn_secondrem}, for
each edge in class $\xi_j$ belonging to $M^\ast_2$, all of the edges in class
$\xi_j$ in matching $G^\ast$ appear in the right hand side with coefficient
$\frac{1}{b_{\xi_j}}$. By definition,there are at most $b_{\xi_j}$ edges of class
$\xi_j$ in $M^\ast$ and exactly $b_{\xi_j}$ edges of this class belong to
$G^\ast$---if this were not the case, Line~\ref{alggreedy_secondremoval} of
Algorithm~\ref{alg:greedy} would not be used. To conclude, the coefficient for
each edge in the right hand side is increased by $\frac{1}{b_{\xi_j}}$ for every
edge in $M^\ast_2 \cap \xi_j$, and summing over all edges, we get a coefficient of one, therefore validating~\eqref{eqn_finalgreedy2}.

Summing~\eqref{eqn_finalgreedy1} and~\eqref{eqn_finalgreedy2}, concludes the
proof. \end{proof}

\subsection{PROOF OF PROPOSITION~\ref{prop:regretdecomp1}}
\label{app:proof:regretdecomp}
\paragraph{Properties of Markov Chains} Before decomposing the regret, we briefly digress to recall some classic results
on mixing of Markov chains. For an ergoidic (i.e.~irreducible and aperiodic) transition matrix on a finite
state space $\Theta$, let $\pi$ be its stationary distribution and $\tilde{P}$
denote the time reversal of its transition matrix $P$---that is,
\[\tilde{P}(\theta,\theta')=\frac{\pi(\theta')P(\theta',\theta)}{\pi(\theta)}.\]
The time reversal kernel $\tilde{P}$ is also ergodic with stationary
distribution $\pi$. Define the \emph{multiplicative reversiblization} $M(P)$ of
$P$ by
$M(P)=P\tilde{P}$
which is a reversible transition matrix itself. The eigenvalues of $M(P)$ are
real and non-negative so that the second largest eigenvalue
$\lambda_1(M)\in[0,1]$~\cite{fill:1991aa}. Define  \emph{chi-squared distance}
from stationary at time $n$ by
\[\chi_n^2=\sum_{\theta}\frac{(\pi_n(\theta)-\pi(\theta))^2}{\pi(\theta)}.\]
where $\pi_n=\sum_{\theta}\pi_0(\theta)P^n(\theta, \cdot)$.

\begin{proposition}[\cite{fill:1991aa}]Let $P$ be an ergodic transition matrix on a finite state space
	$\Theta$ and let $\pi$ be the stationary distribution. 
	Then
	$4\|\pi_n-\pi\|^2\leq \big( \lambda_1(M) \big)^n\chi_0^2$.
	Furthermore,  $\max_{\pi_0\in \mc{P}(\Theta)}\big\|\sum_{\theta}P^n(\theta,\cdot)\pi_0(\theta)-\pi(\cdot)\big\|^2\leq
	\frac{1}{4}\frac{(1-\min_\theta \pi(\theta))^2}{\min_\theta
		\pi(\theta)}\big( \lambda_1(M) \big)^n$. 
	
	where $\mc{P}(\Theta)$ us the space of probability distributions on
	$\Theta$\footnote{We remark that the bound in the above equation is easily computed by noting
		that $\chi_n^2$ is always bounded above by $(\min_\theta
		\pi(\theta))^{-1}(1-\min_\theta \pi(\theta))^2$.}.
	\label{prop:convergence}
\end{proposition}

From the perspective of a general epoch mixing policy $\alpha$, the above proposition provides
a bound on how close the distribution on types for the Markov chain is after
$\tauk$ time steps has elapsed when edge $(a,i)$ is chosen. 

\begin{lemma}
	Consider an arbitrary epoch mixing policy $\alpha$ that selects a matching $\alpha(k)$ during the $k$--th epoch for $\tau_k$ iterations. For each arm $(a,i) \in \alpha(k)$, there exists a constant $C_{a,i}>0$ such that
	\begin{equation}
	\big|\mb{E}\big[ \mu_{a,i}-\bs{r}^{\theta_a(k)}_{a,i}\big]\big|\leq
	\frac{C_{a,i}}{\tauk}
	\label{eq:innerbound}
	\end{equation}
	\label{cor:innerbound}
\end{lemma}
The proof is a direct consequence of Proposition~\ref{prop:convergence}. 
\begin{proof}
	Noting that $\mu^j=\sum_{\theta}r_\theta^j\pi^j(\theta)$, a direct application of Proposition~\ref{prop:convergence} gives us the following:
	\begin{align*}
	&\textstyle    \left|\mb{E}\left[
	\sum_{\theta}r_\theta^j\pi^j(\theta)-\frac{1}{\tauk}\sum_{t=t_k}^{t_{k+1}-1}r_{\theta,t}^{j}\Big|
	\theta_{t_k}\right]\right|\notag\\
	&
	\leq\textstyle
	\frac{1}{\tauk}\sum_{t=t_k}^{t_{k+1}-1}\sum_{\theta}\left|(\pi^j(\theta)-\beta_t(\theta))\right|\\
	&\textstyle\leq
	\frac{1}{\tauk}\sum_{t=t_k}^{t_{k+1}-1}\sum_{\theta}\left|(\pi^j(\theta)-\sum_{\theta'}P_j^{t-t_k}(\theta',\theta)\beta_{t_k}(\theta'))\right|\\
	&\textstyle=
	\frac{1}{\tauk}\sum_{t=t_k}^{t_{k+1}-1}\|\pi^j(\cdot)-\sum_{\theta'}P_j^{t-t_k}(\theta',\cdot)\beta_{t_k}(\theta')\|_{1}\\
	&\textstyle\leq
	\frac{1}{\tauk}\sum_{t=t_k}^{t_{k+1}-1}C_j\lambda_j^{t-t_k}=\frac{C_j(1-\lambda_j^{\tauk})}{\tauk(1-\lambda_j)}
	\end{align*}
	This is simply because of the fact that the expected reward is less than $1$ by
	construction, the triangle
	inequality, 
	and Fubini's
	theorem~\cite[Theorem~2.37]{folland:2007aa}.\end{proof}
We also remark that Proposition~\ref{prop:convergence} also implies that this
bound holds for all $\beta^{(k)}_{a}$ (i.e.~the distribution of agent $a$'s type at the beginning of epoch $k$) and hence, is independent of the algorithm $\alpha$.

\begin{proof}[Proposition~\ref{prop:regretdecomp1}]
Consider the expression for regret from Definition~\ref{defn:regret}:

	\begin{align*}
	\textstyle	R^\alpha(n) & \textstyle=n\sum_{g^\ast_j \in G^\ast}\mu_{g^\ast_j} -
	\sum_{k=1}^n\sum_{(a,i) \in \alpha(k)}\mb{E}[\bs{r}^{\theta}_{a,i}], \end{align*}

By  adding and subtracting $\sum_{(a,i)\in
        \mc{P}}T^\alpha_{a,i}(n)\mu_{a,i}$ from the above equation, the cumulative regret can be written as: 
        \begin{align}
    R^\alpha(n)&= 
    n\sum_{a \in \mc{A}}\mu_{a,i^*(a)} 
    -\sum_{(a,i)\in \mathcal{P}}T^\alpha_{a,i}(n)\mu_{a,i} \nonumber\\ & \quad +\sum_{(a,i)\in
        \mc{P}}T^\alpha_{a,i}(n)\mu_{a,i}-\sum_{k=1}^n\sum_{(a,i) \in
        \alpha(k)}\bs{r}^{\theta_a(k)}_{a,i}\nonumber\\
        &=\sum_{(a,i) \in \mc{P}}T^\alpha_{a,i}(n)\mu_{a,i^*(a)} -\sum_{(a,i)\in \mathcal{P}}T^\alpha_{a,i}(n)\mu_{a,i} \nonumber \\& \qquad +\sum_{(a,i)\in
        	\mc{P}}T^\alpha_{a,i}(n)\mu_{a,i} \nonumber\\& \qquad -\sum_{k=1}^n\sum_{(a,i)\in
            \mc{P}}\mathds{1}\{(a,i) \in\alpha(k)\}\bs{r}^{\theta_a(k)}_{a,i}\nonumber\\	
            &=\sum_{(a,i)\in \mc{P}}T_{a,i}^\alpha(n)(\mu_{a,i^*(a)}-\mu_{a,i})\nonumber\\ & \quad +\sum_{k=1}^n\sum_{(a,i)\in
        \mc{P}}\mathds{1}\{(a,i) \in \alpha(k)\}\big(
        \mu_{a,i}-\bs{r}^{\theta_a(k)}_{a,i} \big) \label{eq:regret1}
         \end{align}
where $\mathds{1}(\cdot)$ is the indicator
 function---e.g.,  $\mathds{1}\{(a,i) \in \alpha(k)\}$ is one when the edge $(a,i)$ belongs to the matching $\alpha(k)$. In the term $\sum_{(a,i) \in \mc{P}}T^\alpha_{a,i}(n)\mu_{a,i^*(a)}$, $\mu_{a, i^*(a)}$ appears exactly $n$ times. 
Although one would expect the matching chosen by the policy (at least in the initial stages) to be sub-optimal compared to the benchmark greedy matching, it is highly possible that some individual edges (arms) may outperform those in the greedy matching. To account for this, we separate the edges in $\mathcal{P}$ into the sub-optimal edges and the super-optimal ones. Formally, for any given $a \in \mc{A}$, define the set of sub-optimal edges $S_a$ as follows:
\[S_a = \{ (a,i) ~|~ \mu_{a,i^*(a)} \geq \mu_{a,i} ~\forall i \in \mc{I}\}.\]

Suppose that $\mc{S} = \bigcup_{a \in \mc{A}} S_a$. Then, the regret bound in Equation~\eqref{eq:regret1} can be simplified by ignoring the contribution of the terms in $\mc{P} \setminus \mc{S}$. That is, since $\mu_{a,i^*(a)} < \mu_{a,i}$ for all $(a,i) \in \mc{P} \setminus \mc{S}$, we have that:
     
\begin{align}
   R^\alpha(n)  & \leq\sum_{(a,i)\in \mc{S}}T_{a,i}^\alpha(n)(\mu_{a,i^*(a)}-\mu_{a,i})  \nonumber \\& \quad  +\sum_{k=1}^n\sum_{(a,i)\in
    	\mc{P}}\mathds{1}\{(a,i) \in \alpha(k)\}\big(
    \mu_{a,i}-\bs{r}^{\theta_a(k)}_{a,i} \big) .
    \label{eqn_regret2_app}
\end{align}

Next, we separate the second term above into the contribution of the edges in $\mc{S}$ and those in $\mc{P} \setminus \mc{S}$. That is,  $\sum_{k=1}^n\sum_{(a,i)\in \mc{P}}\mathds{1}\{(a,i) \in \alpha(k)\}\big(\mu_{a,i}-\bs{r}^{\theta_a(k)}_{a,i} \big)$	can be written as:	
\begin{align}
 \sum_{k=1}^n\sum_{(a,i)\in
   	\mc{S}}\mathds{1}\{(a,i) \in \alpha(k)\}\big(
   \mu_{a,i}-\bs{r}^{\theta_a(k)}_{a,i} \big) \nonumber \\
   +\quad \sum_{k=1}^n\sum_{(a,i)\in
   	\mc{P} \setminus \mc{S}}\mathds{1}\{(a,i) \in \alpha(k)\}\big(
   \mu_{a,i}-\bs{r}^{\theta_a(k)}_{a,i} \big)
    \label{eq:regret2}
\end{align}

We can now use Lemma~\ref{cor:innerbound} to bound the difference between the empirical rewards and the stationary reward during any given epoch. Suppose that $\tau_0\geq 1$ and $\tauk=\tau_0+\zeta k$ with $\zeta$ a non-zero
natural number\footnote{There are other
    choices for the sequence $\{\tauk\}$; e.g., $\tauk=a^k\tau_0$. The choice we
make allows for tighter bounds.}.
An application of Lemma~\ref{cor:innerbound} and the tower property of expectation allows us to bound the first term above, i.e., suppose that $T_1 = \mb{E}\big[\sum_{k=1}^n\sum_{(a,i)\in
	\mc{S}}\mathds{1}\{(a,i) \in \alpha(k)\}\big(
\mu_{a,i}-\bs{r}^{\theta_a(k)}_{a,i} \big)
\big]$. Then,
\begin{align}
   T_1 &=\mb{E}_\alpha\big[\sum_{k=1}^n\sum_{(a,i)\in
    	\mc{S}}\mathds{1}\{(a,i) \in \alpha(k)\}\mb{E}\big[ \nonumber\\
  & \qquad \qquad \mu_{a,i}-\bs{r}^{\theta_a(k)}_{a,i} \big|
       \theta_a(k)    \big]  \big]\nonumber\\
    &\leq \mb{E}_\alpha\big[\sum_{k=1}^n\sum_{(a,i)\in
    	\mc{S}}\mathds{1}\{(a,i) \in \alpha(k)\}\frac{C_{a,i}}{\tauk}
            \big]\nonumber\\
            &\leq \mb{E}_\alpha\big[\sum_{(a,i)\in
            	\mc{S}}\sum_{k=1}^n\mathds{1}\{(a,i) \in \alpha(k)\}\frac{C_{a,i}}{\tau_0}
            \big]\nonumber\\
             &\leq  \sum_{(a,i)\in
             	\mc{S}}\frac{C_{a,i}}{\tau_0}
                \mb{E}_\alpha[T_j^\alpha(n)]
    \label{eq:boundterm1}
\end{align}
where we use the notation $\mb{E}_\alpha$ to emphasize that this
expectation is now dependent only on the algorithm where the number of times an
arm is chosen is a random variable. Analogously, bound the second term of Equation~\ref{eq:regret2}, i.e., $T_2 = \mb{E}\left[\sum_{k=1}^n\sum_{(a,i)\in
	\mc{P} \setminus \mc{S}}\mathds{1}\{(a,i) \in \alpha(k)\}\big(
\mu_{a,i}-\bs{r}^{\theta_a(k)}_{a,i} \big)    \right]$
\begin{align*}
    T_2    &\leq \mb{E}_\alpha\big[\sum_{k=1}^n\sum_{(a,i)\in
    	\mc{P} \setminus \mc{S}}\mathds{1}\{(a,i) \in \alpha(k)\}\frac{C_{a,i}}{\tauk}
    \big]\\
        &\leq \sum_{k=1}^n\frac{1}{\tauk}\sum_{(a,i)\in
        	\mc{P}\setminus \mc{S}}C_{a,i} \mb{E}_\alpha\big[\mathds{1}\{(a,i) \in \alpha(k)\}
        \big]\\
       &\leq C_{\ast}\sum_{k=1}^n\frac{1}{\tauk}\sum_{(a,i)\in
       	\mc{P}} \mb{E}_\alpha\big[\mathds{1}\{(a,i) \in \alpha(k)\}
       \big]\\
&\leq  mC_{\ast}\sum_{k=1}^n\frac{1}{\tauk},
\end{align*}

where $C_{\ast} = \max_{(a,i) \in \mc{P} \setminus \mc{S}}C_{a,i}$. Note that for any given epoch $k$, our policy selects at most $m$ edges in the matching and therefore, $\sum_{(a,i)\in \mc{P}\setminus L}\mb{E}_\alpha\big[\mathds{1}\{(a,i) \in \alpha(k)\}\big] \leq \sum_{(a,i)\in \mc{P}}\mb{E}_\alpha\big[\mathds{1}\{(a,i) \in \alpha(k)\}\big] \leq m$. Finally, we can bound the harmonic summation using the fact that $\tau_k = \tau_0 + \zeta k$:
\begin{align}
T_2 &\leq mC_{\ast}\sum_{k=1}^n\frac{1}{\tauk} \nonumber\\
& \leq mC_{\ast} \frac{1}{\zeta}\Big(
1+\int_{\tau_0/\zeta}^{n-1+\tau_0/\zeta}\frac{1}{x}dx\Big)\nonumber\\
& \leq mC_{\ast}\frac{1}{\zeta}\Big(
    1+\log\Big( \frac{n-1}{\tau_0}+1 \Big)\Big)\label{eq:boundterm2}
\end{align}
Recall from the definition of the marginal infeasibility sets in Equation~\eqref{eqn_infeasibilityset} that for any given $(a,i) \in \mc{P} \setminus \mathcal{G^*}$, there exists a unique edge $g^*_j \in G^*$ such that $(a,i) \in L^*_j$. Define $L^{-1}(a,i) \coloneqq g^*_j \in G^*$ such that $(a,i) \in L^*_j$. Now, we can define the reward gap for any given edge $(a,i) \in \mc{P}$ as follows: 
\begin{align*}
\Delta_{a,i} & =\mu_{a,i^*(a)}- \mu_{a,i} \qquad  \text{if $(a,i) \in \mc{S}$}\\
& = \mu_{L^{-1}(a,i)} - \mu_{a,i} \quad ~ \text{if $(a,i) \in (\mc{P} \setminus G^*) \setminus \mc{S}$}\\
& = \mu_{g^*_{j-1}} - \mu_{g^*_j} \qquad ~~\text{if $(a,i) = g^*_j$ for $j \geq 2$}.
\end{align*}

Going back to our regret lower bound in~\eqref{eqn_regret2_app} and decomposing the second term using~\eqref{eq:boundterm1} and~\eqref{eq:boundterm2}, we get the main proposition. \end{proof}  

\subsection{PROOF OF THEOREM~\ref{thm:regretbound}}
\label{app:proof:regretbound}
Before proving Theorem~\ref{thm:regretbound}, we state some useful supplementary lemmas. 

	\begin{lemma}[Azuma-Hoeffding Inequality~\cite{azuma:1967aa,hoeffding:1963aa}]
		Suppose $(Z^k)_{k\in \mb{Z}_+}$ is a martingale with respect to the
		filtration $(\mc{F}^k)_{k\in \mb{Z}_+}$ having bounded differences, i.e.,
		there are finite, non-negative constants $c^k$, $k\geq 1$ such that
		$|Z^k-Z^{k-1}|<c^k$ almost surely. Then for all $t>0$ 
		\begin{equation*}
		P(Z^k-\mb{E}Z^k\leq -t)\leq \exp\left(
		-\frac{t^2}{2\sum_{k=1}^N(c^k)^2} \right).
		\end{equation*}
		\label{prop:AH}
	\end{lemma}

We define some notation that is useful for the following lemma as well the proof of Theorem~\ref{thm:regretbound}. Consider the MG-EUCB algorithm described in Algorithm~\ref{alg:ucb}. Let $R^{\theta,j}_{a,i}$ be the
	cumulative reward received when arm $(a,i)$ is chosen for the $j$--th time
	where we
	include $\theta$ in the subscript to note the state-dependence of the random
	reward. 
	That is,
	$R^{\theta,j}_{a,i}=\bs{r}_{a,i}^{\theta_a(t^j_{a,i})}$ where, by an abuse
	of notation, $t_{a,i}^j$ denotes the time instance at which edge $(a,i)$ is pulled for the $j$--th time and $\theta_a(t^j_{a,i})$ denotes the state of agent $a$ during that epoch.
	
	Define the filtration $\mc{F}^k_{a,i}=\sigma(R^{\theta,1}_{a,i}, \ldots, R^{\theta,
		k}_{a,i},\theta_a(t_1^j), \ldots,
	\theta_a(t_k^j))$---that is, the smallest
	$\sigma$-algebra generated by the random variables $(R^{\theta,1}_{a,i}, \ldots, R^{\theta,
		k}_{a,i},\theta_a(t^1_{a,i}), \ldots, \theta_a(t^k_{a,i}))$. Let
	$X_{a,i}^k=R^{\theta,k}_{a,i}-\mb{E}[R^{\theta,k}_{a,i}|\mc{F}^{k-1}_{a,i}]$ and $Y_{a,i}^k=\sum_{j=1}^kX_{a,i}^j$. 
	We have that $Y_{a,i}^k$ is a martingale since
	$\mb{E}[Y^{k+1}_{a,i}|\mc{F}_{a,i}^k]=\mb{E}[X^{k+1}_{a,i}|\mc{F}_{a,i}^k]+\mb{E}[Y_{a,i}^k|\mc{F}_{a,i}^k]=Y_{a,i}^k$
	(since $Y_{a,i}^k$ is $\mc{F}_{a,i}^k$--measurable by construction) and
	$\mb{E}[|Y_{a,i}^k|]<\infty$ (rewards are bounded). Moreover, the boundedness of the
	rewards also implies the martingale $Y_{a,i}^k$
	has bounded differences. Indeed, $|Y_{a,i}^k-Y^{k-1}_{a,i}|=|X_{a,i}^k|\leq 1$ almost surely since rewards
	are normalized to be on the interval $[0,1]$, without loss of generality. 	Now, we are ready to show an upper bound on the difference in the empirical reward and the stationary state rewards. 

\allowdisplaybreaks

		\begin{lemma} Given aperiodic, irreducible Markov chains $P_{a,i}$ with
			corresponding stationary distributions $\mu_{a,i}$ for each $(a,i)\in \mc{P}$ and mixing
			sequence $\{\tau_k\}$ such that $\tau_k=\tau_0+\zeta k$, $\tau_0\geq1$, we have
			that
			\begin{align}
			&\textstyle\left|\mb{E}\left[\mu_{a,i}-\frac{1}{k}\sum_{j=1}^{k}\mb{E}[R^{\theta,j}_{a,i}|\mc{F}^{j-1}_{a,i}]\right]\right|\nonumber\\
            &\leq
            \textstyle
			\frac{C_{a,i}}{2k}\left(\frac{1}{\zeta+\tau_0}+\frac{1}{\zeta}\log\left(
			1+\frac{k\zeta}{\tau_0} \right) \right)
			\label{eq:lemmabound}\end{align}
			\label{lem:boundforAH}
		\end{lemma} 
		The proof of the above lemma follows a similar line of reasoning as Lemma~\ref{cor:innerbound}. 
		\begin{proof}
			Since $\Theta$ is a finite set with finite
			elements (i.e.~$|x|<\infty$ for all $x\in \Theta$), we are able to use analogous
			reasoning as was used in Proposition~\ref{prop:convergence} along with the
			Markov property on the conditional expectation $\mb{E}[R_i^j|\mc{F}_{i-1}^j]$ to bound
			$\mu^j-\frac{1}{k}\sum_{i=1}^k\mb{E}[R_i^j|\mc{F}_{i-1}^j]$ by
			$\frac{L_j(k)}{k}$ for some constant $L_j(k)$. Indeed, the quantity $V = \Big|\mb{E}\big[\mu^j-\frac{1}{k}\sum_{i=1}^{k}\mb{E}[R_i^j|\mc{F}_{i-1}^j]\big]\Big|$ can be simplified as follows:
			\begin{align*}
			\textstyle
			V &=\textstyle\left|\frac{1}{k}\sum_{i=1}^k\left(\mb{E}\left[\mu^j-\mb{E}[R_i^j|\mc{F}_{i-1}^j]\right]\right)\right|\\
			&\textstyle\leq
			\frac{1}{k}\sum_{i=1}^{k}\mb{E}\Big[\sum_{\theta}r_{\theta}^j\pi^j(\theta)- \\
			& \quad \mb{E}\big[(\tau_i^j)^{-1}\sum_{t=t_i^j}^{t_{i+1}^j-1}r_{\theta,t}^j\Big|\mc{F}_{i-1}^j\big]\Big]\\
			&\textstyle\leq \frac{1}{k}\sum_{i=1}^{k}\mb{E}\Big[
			(\tau_i^j)^{-1}\sum_{t=t_i^j}^{t_{i+1}^j-1}\sum_{\theta}|\pi^j(\theta)-\beta_t(\theta)|
			\Big]\\
			&\textstyle\leq
			\frac{1}{k}\sum_{i=1}^{k}\frac{C_j}{2}\mb{E}\Big[(\tau_i^j)^{-1}\sum_{t=t_i^j}^{t_{i+1}^j-1}(\lambda_j)^{(t-t_i^j)}\Big]\\
			&\textstyle\leq
			\frac{1}{k}\sum_{i=1}^{k}\frac{C_j}{2}\mb{E}\left[(\tau_i^j)^{-1}(1-(\lambda_j)^{\tau_i^j})(1-\lambda_j)^{-1}\right]
			\\
			&\textstyle\leq
			\frac{1}{k}\frac{C_j}{2}\frac{1}{1-\lambda_j}\sum_{i=1}^{k}\mb{E}\big[(\tau_i^j)^{-1}\big],
			\end{align*}
			where we have used the fact that the reward bounded almost surely on $[0,1]$. 
			Now, $1/\tau_{i^j}$ is a random variable with respect to the algorithm since at
			the $i$--th pull of arm $j$ we do not know \emph{a priori} what iteration of the
			algorithm we are on. However, at the $i$--th pull of arm, we do know that the
			algorithm is at least at the $i$--th iteration. Hence, 
			$\sum_{i=1}^{k}\mb{E}\big[(\tau_i^j)^{-1}\big]\leq
			\sum_{i=1}^k(\tau_0+\zeta i)^{-1}$.
			Now, 	for any $a \geq 1$ and positive integer $k$, we have that
			$\sum_{i=a}^{a+k}(i)^{-1} \leq \frac{1}{a} + \log(1 + \frac{k}{a}).$
			Indeed, rewrite the summation in the lemma statement as 
			$\sum_{i=a}^{a+k}i^{-1} = a^{-1} + \sum_{i=a+1}^{a+k}i^{-1}$ and apply 
			the fundamental inequality, 
			$(i)^{-1} \leq \int_{i-1}^i x^{-1}dx$,
			which holds for any $i \geq 1$, repeatedly  for $i=a+1, a+2,
			\ldots, a+k$ so that we have  a telescoping summation of integrals---i.e.
			\begin{align*}
			\textstyle\sum_{i=a}^{a+k}\frac{1}{i} & = \textstyle\frac{1}{a} + \sum_{i=a+1}^{a+k}\frac{1}{i} \\ &\leq
			\frac{1}{a} + \int_{a}^{a+k}\frac{1}{x}dx  = \frac{1}{a} +
			\log\left(\frac{a+k}{a}\right).
			\end{align*}
			Thus,
			$\sum_{i=1}^k(\tau_0+\zeta i)^{-1}\leq
			(\zeta+\tau_0)^{-1}+\frac{1}{\zeta}\log\big(
			1+\frac{k\zeta}{\tau_0} \big)$
			so that \eqref{eq:lemmabound} holds.
		\end{proof}

    \begin{proof}[Theorem~\ref{thm:regretbound}]
We begin by formalizing the choice of the UCB  parameter $c^k_{a,i}(t)$---it is crucial that this parameter reflects the error due to both the Markov chain and the randomness of rewards. Applying Lemma~\ref{lem:boundforAH} to our problem, we observe that the average error stemming from the randomness in the user state after $k$ pulls of the edge $(a,i)$ can be written as:
\begin{align*}
&\textstyle\left|\mb{E}\left[\mu_{a,i}-\frac{1}{k}\sum_{j=1}^{k}\mb{E}[R^{\theta,j}_{a,i}|\mc{F}^{j-1}_{a,i}]\right]\right|\notag\\
&\leq
\textstyle
\frac{C_{a,i}}{2k}\left(\frac{1}{\zeta+\tau_0}+\frac{1}{\zeta}\log\left(
1+\frac{k\zeta}{\tau_0} \right) \right)
\end{align*}
Based on this, for each edge $(a,i)$ and `pull count' $k$, we define the constant $Q_{a,i}(k)$
	\begin{align*}
	Q_{a,i}(k)=\frac{C_{a,i}}{2}\left(\frac{1}{\zeta+\tau_0}+\frac{1}{\zeta}\log\left(
	1+\frac{k\zeta}{\tau_0} \right) \right).
	\end{align*}
	Finally, we can now define the confidence parameter as follows: 
\[c^k_{a,i}(t)=Q_{a,i}(k)/k+\sqrt{\frac{6}{k}\log(t) + \frac{4}{k} \log(m) }.\]
		
	Coming back to the proof of Theorem~\ref{thm:regretbound}, our primary goal is to map every selection of a sub-optimal edge to a condition on the relative empirical rewards between edges that can then be resolved using Azuma-Hoeffding inequality. 	
Applying Lemma~\ref{lem_greedyordering_main}, we see that if
    MATCHGREEDY does not return the benchmark matching $G^\ast$ at epoch $t$ and
    instead returns a matching $\alpha(t) \neq G^\ast$, at least one of the
    above conditions must fail. Alternatively, this implies that one of the
    following two (inverse) conditions must be true:
	\begin{enumerate}[itemsep=0pt, topsep=0pt, leftmargin=15pt]
\item $\mathds{1}\{\exists j < j'|\ \big(u_{g^\ast_{j'}}(t) >
    u_{g^\ast_j}(t)\big) \land ( g^\ast_{j'} \in \alpha(t))\}$
\item $\mathds{1}\{\exists j, (a,i) \in L^\ast_j|\ \big( u_{g^*_{j}}(t) <
    u_{a,i}(t)\big) \vee ((a,i) \in \alpha(t))    \} =1$
	\end{enumerate}
		To express the above conditions in a concise manner, let us augment the sets
    $L^\ast_j$ to include edges from the greedy matching. Specifically, for all $1
    \leq j \leq m-1$, let $L^{+}_j = L^\ast_j \cup \{g^\ast_{j+1}\}$ and $L^+_m
    = L^\ast_m$. Observe that $\bigcup_{j} L^+_j = \mc{P} \setminus g^\ast_1$.  Now, we can formally say that if the matching returned by the UCB algorithm during iteration $t$ (call this matching $\alpha(t)$) does not coincide with the greedy matching, then
	\begin{align}
	\label{eqn_matchingcharacterization}
&	\mathds{1}\{\exists 1\leq j\leq m, (a,i) \in L^+_j|\ u_{g^\ast_j}(t) \notag\\
&\quad< u_{a,i}(t) \land (a,i) \in \alpha(t)\} = 1.
	\end{align}
	
   We will use
    the notation $\bar{R}^k_{a,i}=\frac{1}{k}\sum_{j=1}^k R^{\theta,j}_{a,i}$.  Since Proposition~\ref{prop:regretdecomp1} provides an upper bound for the regret in terms of the number of times each (sub-optimal) edge is chosen, it suffices to bound the quantity $T_{a',i'}(n)$, which is the number of times our UCB algorithm selects the edge $(a',i')$ given that $(a',i') \in \mc{S}$---i.e. $\mu_{a',i'} < \mu_{a', i^*(a')}$. Note that by definition, for any $(a,i) \in \mc{S}$, the edge $(a',i')$ does not belong to the greedy benchmark matching $G^*$.  Suppose that $\ell$ denotes an arbitrary integer (to be formalized later). Then, we have that: \begin{align*}
    T_{a',i'}&(n)=\textstyle 1+\sum_{t=m+1}^n\mathds{1}\{(a',i') \in \alpha(t)\}\\
    &\textstyle\leq 1 +\sum_{t=m+1}^n 	\mathds{1}\{\exists j, (a,i) \in L^+_j|\
        u_{g^\ast_j}(t)\\
        &\quad< u_{a,i}(t) \land (a,i) \in \alpha(t)\} \quad (\text{from}~\eqref{eqn_matchingcharacterization}) \\   
    &\leq \textstyle 1 +\sum_{t=m+1}^n \sum_{j=1}^m \sum_{(a,i) \in L^+_j}
    \mathds{1}\{ u_{g^\ast_j}(t) \leq \\
    & \quad    u_{a,i}(t) \land (a,i) \in \alpha(t)\} \quad 
    \\ 
    &= \textstyle 1 +\sum_{j=1}^m \sum_{(a,i) \in L^+_j} \sum_{t=m+1}^n
    \mathds{1}\{ u_{g^\ast_j}(t)\\
    &\quad\leq u_{a,i}(t) \land (a,i) \in \alpha(t)\} \\ 
    &\leq \textstyle  1 +\sum_{j=1}^m \sum_{(a,i) \in L^+_j}\big(\ell\\
    &\quad\textstyle+ \sum_{t=m+1}^n  \mathds{1}\{ u_{g^*_j}(t) \leq u_{a,i}(t)\\
    &\qquad\land
    (a,i) \in \alpha(t) \land T_{a,i}(t) > \ell\} \big)\\
        &\leq \textstyle  1 +\sum_{j=1}^m \sum_{(a,i) \in L^+_j}\big(\ell\\
        &\quad\textstyle+
        \sum_{t=m+1}^n  \mathds{1}\{ u_{g^*_j}(t) \leq u_{a,i}(t) \\
        &\qquad\land
        T_{a,i}(t) > \ell\} \big)\\
    &\leq\textstyle  \ell m^2 +\sum_{j=1}^m \sum_{(a,i) \in L^+_j}\sum_{t=m+1}^n
    \mathds{1}\{ u_{g^*_j}(t)\\
    &\quad\leq u_{a,i}(t) \land  T_{a,i}(t) > \ell\}\\
    &\leq \textstyle \ell m^2+\sum_{j=1}^m \sum_{(a,i) \in L^+_j}\sum_{t=m+1}^n\big(\\
    &\quad\textstyle \mathds{1}\{\min_{0<s<t}u^s_{g^*_j}(t) \leq  \max_{\ell\leq
    k<t}  u^k_{a,i}(t) \}\big)\\
    &\leq \textstyle \ell m^2 +\sum_{j=1}^m \sum_{(a,i) \in L^+_j}\big(\\
    &\quad\textstyle\sum_{t=m+1}^n \sum_{s=1}^{t-1}\sum_{k=\ell}^{t-1}
    \mathds{1}\{u^s_{g^*_j}(t)  \leq    u^k_{a,i}(t) \}\big)\\
    & = \textstyle \ell m^2 +\sum_{j=1}^m
    \sum_{(a,i) \in L^+_j}\sum_{t=m+1}^n\sum_{s=1}^{t-1}\big(\\
    &\quad\textstyle\sum_{k=\ell}^{t-1}
    \mathds{1}\{\bar{R}^s_{g^*_j} + c^s_{g^*_j}(t) \leq   \bar{R}^k_{a,i} +
    c^k_{a,i}(t)  \}\big)
    \end{align*}
    Now, $\bar{R}^s_{g^\ast_j} + c^s_{g^\ast_j}(t) \leq   \bar{R}^k_{a,i} + c^k_{a,i}(t)$ implies that atleast one of the following
    must hold:
    \begin{eqnarray}
    \label{eq:cond1}
    \bar{R}^s_{g^*_j} &\leq&\mu_{g^*_j} -  c^s_{g^*_j}(t) \\
    \label{eq:cond2}
    \bar{R}^k_{a,i}&\geq & \mu_{a,i}+c^k_{a,i}(t)\\
    \mu_{g^*_j}  &<&\mu_{a,i}+ 2c^k_{a,i}(t)
    \label{eq:cond3}
    \end{eqnarray}
 Indeed, suppose that all three of the above inequalities are false. Then, 
$u^s_{g^\ast_j}(t) = \bar{R}^s_{g^\ast_j}+c^s_{g^\ast_j}(t)>\mu_{g^\ast_j} \geq
        \mu_{a,i}+2c^k_{a,i}(t)>\bar{R}^k_{a,i}+c^k_{a,i}(t) = u^k_{a,i}(t)$, which is, of course, a contradiction.
Hence, if
    $\bar{R}^s_{g^\ast_j}+c^s_{g^\ast_j}(t)\leq
    \bar{R}^k_{a,i}+c^k_{a,i}(t)$, then at least one of
    \eqref{eq:cond1}--\eqref{eq:cond3} holds.        
                We bound the probability of events \eqref{eq:cond1} and \eqref{eq:cond2}
        using the Azuma-Hoeffding inequality in Lemma~\ref{prop:AH} and
        find an $\ell$ such that \eqref{eq:cond3} is always false for every $j, (a,i) \in L^+_j$.

    Towards this end, we apply Lemma~\ref{prop:AH} to the martingale $(Y^k_{a,i})_{k\in
    \mb{Z}_+}$. Note that by the law of conditional expectations,
    $\mb{E}[Y^k_{a,i}]=0$ so that Lemma~\ref{prop:AH} implies that for each
    arm $(a,i)$ and any $t>0$,
        $P(Y^k_{a,i}\leq -t)\leq \exp(-t^2/(2k) )$.

We need to relate the random variable $Y^k_{a,i}$ to 
the difference of the empirical mean of the
average cumulative reward from its true value for each arm so that we can bound
this difference. Consider the event
\begin{align*}
  \omega&=\big\{ \mu_{g^*_j}-\bar{R}^s_{g^*_j}\geq \gamma \big\}\\
    &=\textstyle\big\{ \mu_{g^*_j}-\frac{1}{s}\sum_{l=1}^s
    \mb{E}[R^{\theta,l}_{g^*_j}|\mc{F}^{l-1}_{g^*_j}]\\
    &\quad \textstyle+\frac{1}{s}\sum_{l=1}^{s}
    \mb{E}[R^{\theta,l}_{g^*_j}|\mc{F}^{l-1}_{g^*_j}]-\bar{R}^s_{g^*_j}\geq
\gamma \big\}\\
    &=\textstyle\big\{ \mu_{g^*_j}- \frac{1}{s}\sum_{l=1}^s
    \mb{E}[R^{\theta,l}_{g^*_j}|\mc{F}^{l-1}_{g^*_j}]-\frac{1}{s}Y^s_{g^*_j}\geq
\gamma\big\}
\end{align*}
where we have added and subtracted the random variable $\frac{1}{s}\sum_{l=1}^s
\mb{E}[R^{\theta,l}_{g^*_j}|\mc{F}^{l-1}_{g^*_j}]$.
By Lemma~\ref{lem:boundforAH}, 
\begin{align*}
 \omega &\subset \textstyle\big\{ \frac{1}{s}Q_{g^*_j}(s)
    -\frac{1}{s}Y^s_{g^*_j}\geq \gamma \big\}\\
    &=\textstyle \big\{ \frac{1}{s}Y^s_{g^*_j}\leq
\frac{1}{s}Q_{g^*_j}(s)-\gamma \big\}.
\end{align*}
Hence,
\begin{align*}    P\big( \mu_{g^*_j}- \bar{R}^s_{g^*_j} \geq \gamma\big)&\leq
    P\Big( \frac{1}{s}Y^s_{g^*_j}\leq
    \frac{1}{s
    }Q_{g^*_j}(s)-\gamma
    \Big)\\
    &\leq 
\exp\big( -\frac{1}{2}s\big(\gamma-\frac{1}{s}Q_{g^*_j}(s)
\big)^2\big)\end{align*}
so that with 
$\gamma=c^s_{g^*_j}(t) = \sqrt{\frac{6}{s}\log t + \frac{4}{s}\log m}+\frac{1}{s}Q_{g^*_j}(s)$,
we have, 
\[    P\left(\mu_{g^*_j}- \bar{R}^s_{g^*_j}\geq c^s_{g^*_j}(t) 
 \right)\leq t^{-3} m^{-2}.\]
Therefore, it follows that
$P(\bar{R}^s_{g^*_j}\leq \mu_{g^*_j}-c^s_{g^*_j}(t))\leq t^{-3} m^{-2}$ and
  $P(\bar{R}^k_{a,i}\geq \mu_{a,i}+c^k_{a,i}(t)^\ast)\leq t^{-3}m^{-2}$
which imply that \eqref{eq:cond1} and \eqref{eq:cond2} occur with very low probability.

Now, we choose $\ell$ to be the largest integer such that \eqref{eq:cond3} is always
false.  
Indeed, we choose it such that
\begin{align*}
    &\mu_{g^*_j}-\mu_{a,i}-2c^k_{a,i}(t)\\
    &>\textstyle\mu_{g^*_j}-\mu_{a,i}-2\left(
\frac{Q_{a,i}(\ell)}{\ell}+\sqrt{\frac{6\log t}{\ell} + \frac{4\log m}{\ell}} \right)>0.
\end{align*}
Plugging in $Q_{a,i}(\ell)$, we have
\begin{align}
   &\textstyle \Delta_{a,i}-2\Big(
\frac{C_{a,i}}{2\ell}\left(\frac{1}{\zeta+\tau_0}+\frac{1}{\zeta}\log\left(
    1+\frac{\ell\zeta}{\tau_0} \right) \right)\notag\\
    &\textstyle\quad+\sqrt{\frac{1}{\ell}6\log t + \frac{1}{\ell}4\log m}
\Big)>0.
\label{eq:ineq11}\end{align}

Let $\tilde{\ell}=\ell\zeta/\tau_0$ so that
\begin{align*}   \textstyle \Delta_{a,i}-2\Big(
   & \textstyle\frac{C_{a,i}}{2\tau_0}\left(\frac{1}{\tilde{\ell}}\frac{\zeta}{\zeta+\tau_0}+\frac{1}{\tilde{\ell}}\log\left(
    1+\tilde{\ell} \right) \right)\\
    &\textstyle+\sqrt{\frac{6\log t}{\ell} + \frac{4\log m}{\ell}}
    \Big)>0.\end{align*}
Since $1/x<1/\sqrt{x}$ and $1/x\log(1+x)<1/\sqrt{x}$ on $[1,\infty)$, we
    have that
    \[\frac{1}{\tilde{\ell}}\frac{\zeta}{\zeta+\tau_0}+\frac{1}{\tilde{\ell}}\log\left(
        1+\tilde{\ell}
    \right)<\frac{\zeta}{\zeta+\tau_0}\frac{1}{\sqrt{\tilde{\ell}}}+\frac{1}{\sqrt{\tilde{\ell}}}\]
so that \eqref{eq:ineq11} reduces to finding the largest integer $\ell$ such that
\begin{align*}
\Delta_{a,i}-2\bigg( \frac{C_{a,i}}{2\tau_0}\left(
    \frac{\zeta}{\zeta+\tau_0}\frac{\sqrt{\tau_0}}{\sqrt{\ell\zeta}}+\frac{\sqrt{\tau_0}}{\sqrt{\ell\zeta}}
\right) \\ +\frac{\sqrt{6\log t + 4\log m}}{\sqrt{\ell}} \bigg)>0
\end{align*}
Rearranging and squaring terms, we get that~\eqref{eq:cond3} is false for
    \begin{equation}
        \ell \geq \Big\lceil \frac{4}{\Delta_{a,i}^2}\big(\frac{\rho_{a,i}}{\sqrt{\tau_0}}
+\sqrt{6\log n + 4\log m} \big)^2\Big\rceil.
        \label{eq:ellchoice4}
   \end{equation}
  In the above equation, $\rho_{a,i}$ is the edge-specific constant
$$\rho_{a,i}= (
    \frac{\zeta}{\zeta+\tau_0}+1 )\frac{C_{a,i}}{2\sqrt{\zeta}}.$$
In fact, we require that \eqref{eq:cond3}  be false for all $1 \leq j\leq m$ and $(a,i) \in L^+_j$.  Therefore, we set the parameter $\ell$ to be the maximum of the right hand side of~\eqref{eq:ellchoice4}. Formally, define $(a^*, i^*)$ to be the edge in $\mc{P} \setminus g^*_1$ that maximizes the right hand side of~\eqref{eq:ellchoice4}. That is, for a given instance, 

\begin{align}
(a^*,i^*) &=  \argmax\limits_{(a_1, i_1) \in \mc{P} \setminus g^*_1}  \Bigg\lceil \frac{4}{\Delta_{a_1,i_1}^2}\bigg( \frac{\rho_{a_1,i_1}}{\sqrt{\tau_0}} \nonumber \\ & +\sqrt{6\log n + 4\log m} \bigg)^2\Bigg\rceil\label{eqn_astaristar}
\end{align}
Then, by defining $\ell$ as follows, we are assured that Equation~\ref{eq:ellchoice4} holds for all $1 \leq j \leq m$ and $(a,i) \in L^+_j$. 
\begin{equation}
\ell=    
\Bigg\lceil \frac{4}{\Delta_{a^*, i^*}^2}\left( \frac{\rho_{a^*, i^*}}{\sqrt{\tau_0}}+\sqrt{6\log n + 4\log m} \right)^2\Bigg\rceil
\label{eq:ellchoice3}
\end{equation}

Hence, we can bound the number of plays of our original sub-optimal arm $(a',j')$ as follows:
\begin{align*}
    \mb{E}[T_{a',i'}(n)]&\leq \ell m^2 +
    \sum_{j=1}^m \sum_{(a,i) \in L^+_j}\sum_{t=m+1}^n\\&\quad\sum_{s=1}^{t-1}\sum_{k=\ell}^{t-1}\big(
    P(\bar{R}^s_{g^*_j}\leq \mu_{g^*_j}-c^s_{g^*_j}(t)) \\&\quad+P(\bar{R}^{k}_{a,i}\geq
    \mu_{a,i}+c^k_{a,i}(t))\big)\\
    &\leq \Big\lceil \frac{4m^2}{\Delta_{a^*,i^*}^2}\Big(
    \frac{\rho_{a^*,i^*}}{\sqrt{\tau_0}}+\sqrt{6\log n + 4\log m} \Big)^2\Big\rceil
 \\& \quad + 
       \sum_{(a,i) \in \mc{P}}\sum_{t=1}^n\sum_{s=1}^{t}\sum_{k=1}^{t}2t^{-3}m^{-2}\\
       &\leq \frac{4m^2}{\Delta_{a^*,i^*}^2}\Big(
    \frac{\rho_{a^*,i^*}}{\sqrt{\tau_0}}+\sqrt{6\log n + 4\log m} \Big)^2\\&\quad+2(1+\log( n)).
    \label{eq:Tbound}
\end{align*}\end{proof}
As a direct consequence of Theorem~\ref{thm:regretbound}, we can bound the regret of the MatchGreedy-EpochUCB policy.
\begin{corollary}[Regret Bound for UCB]
    Consider $\alpha$ as the MatchGreedy-EpochUCB algorithm and suppose that $\tau_k=\tau_0+\zeta k$ with $\tau_0\geq 1$. The regret bound is
    \begin{align*}
     R^{\alpha}(n) &\leq   \sum_{(a,i) \in \mc{S}}\Bigg( \frac{4m^2}{\Delta_{a^*,i^*}^2}\big(
        \frac{\rho_{a^*,i^*}}{\sqrt{\tau_0}} +\sqrt{6\log n + 4\log m} \big)^2 \nonumber\\ &\quad +2(1 +\log
        (n))\Bigg)\left(\Delta_{a,i}+\frac{C_{a,i}}{\tau_0}\right) \qquad \nonumber\\ & \qquad +\frac{mC_{\ast}}{\zeta}\left(
    1+\log\left( \frac{\zeta(n-1)}{\tau_0}+1 \right)\right),
\end{align*}
where $(a^*,i^*)$ is an edge defined in~\eqref{eqn_astaristar} and $\rho_{a^*, i^*}$ and $C_{a,i}$ are edge-specific constants.
\label{corr:ucb}
\end{corollary}

\section{UCB ALGORITHM}
\label{app:algs}
\subsection{INITIAL PLAY OF UCB ALGORITHM}
\label{app:permutematchings}
Since the UCB algorithm estimates the average reward for each edge $(a,i)$, it is customary to initialize a preliminary round where each arm is played exactly once. In the absence of any capacity constraints (e.g., $b_{\xi_l} = m$ for all $\xi_l \in \mc{C}$), it is easy to compute a sequence of $m$ matchings so that every edge in $\mc{P}$ belongs to exactly one of these matchings. We now present a procedure that achieves the same effect even in the presence of arbitrary capacity constraints.

\begin{algorithm}
	\caption{Computation of disjoint matchings that play each arm once}
	\label{alg:initial}
	\begin{algorithmic}[1] 
		\Function{Matchings-InitialPlay}{$\mc{P}$} 
		\State $\mc{E} \gets \mc{P}$ \Comment{Edges not yet selected}
		\State $i \gets 1$ \Comment{Index for current matching}
		\While{$\mc{E} \neq \emptyset$}
		\State $F \gets \mc{E}$ \Comment{Feasible set for current matching}
		\State $M \gets \emptyset$
		\While{$F \neq \emptyset$}
		\State Select any $(a,i) \in F$
		\If{$M \cup (a,i)$ does not violate~\eqref{eq:lp_matching}}
		\State $M \gets M \cup (a,i)$
		\Else
		\State $F \gets F \setminus (a,i)$.
		\EndIf
	\EndWhile
	\State $M_i \gets M$, $i \gets i+1$, $\mc{E} \gets \mc{E} \setminus M$.
		\EndWhile
		\State\Return $M_1, M_2, \ldots, M_{i-1}$
		\EndFunction
	\end{algorithmic}
\end{algorithm}

Informally, in some iteration $i$, the above algorithm greedily selects edges for matching $M_i$ without violating the capacity constraints. When no additional edge can be added to $M_i$---a maximal matching---we move on to the next iteration. 

Unfortunately, the number of matchings returned by this procedure can be quite large---in the worst case this can be as large as $m^2$, where $m$ is the number of agents or incentives. However, for more reasonable instances such as the ones considered in our simulations, we observe that the number of initial matchings required to play each edge at least once is much closer to the lower bound of $m$.

\begin{algorithm}[h]
    \caption{Environment Implementation for Pulling a Matching (Set of Arms)}
    \label{alg:armpull}
    \begin{algorithmic}[1] 
                \Function{incent}{$M$, $t_n$, $n$ $\tau_0$,$\zeta$} 	       \State $r_{a,i}^{t_n}\gets 0$ ~ $\forall (a,i) \in M$
                 \For{$t \in [t_n,t_n+\tau_0+\zeta n-1]$} 
                 \For{$(a,i) \in M$}
            \State offer incentive $i$ to agent $a$
            \State receive reward $r^{\theta_a,t}_{a,i}$ 
            \State $r^{t_n}_{a,i}\gets r^{\theta,t}_{a,i} + r^{t_n}_{a,i}$
            \EndFor
            \EndFor
            \State\Return $({r}^{t_n}_{a,i})_{(a,i) \in M}$
           \EndFunction

    \end{algorithmic}
\end{algorithm}

\begin{figure*}[t]
    \centering
    \subfloat[]{\includegraphics[width=.5\textwidth]{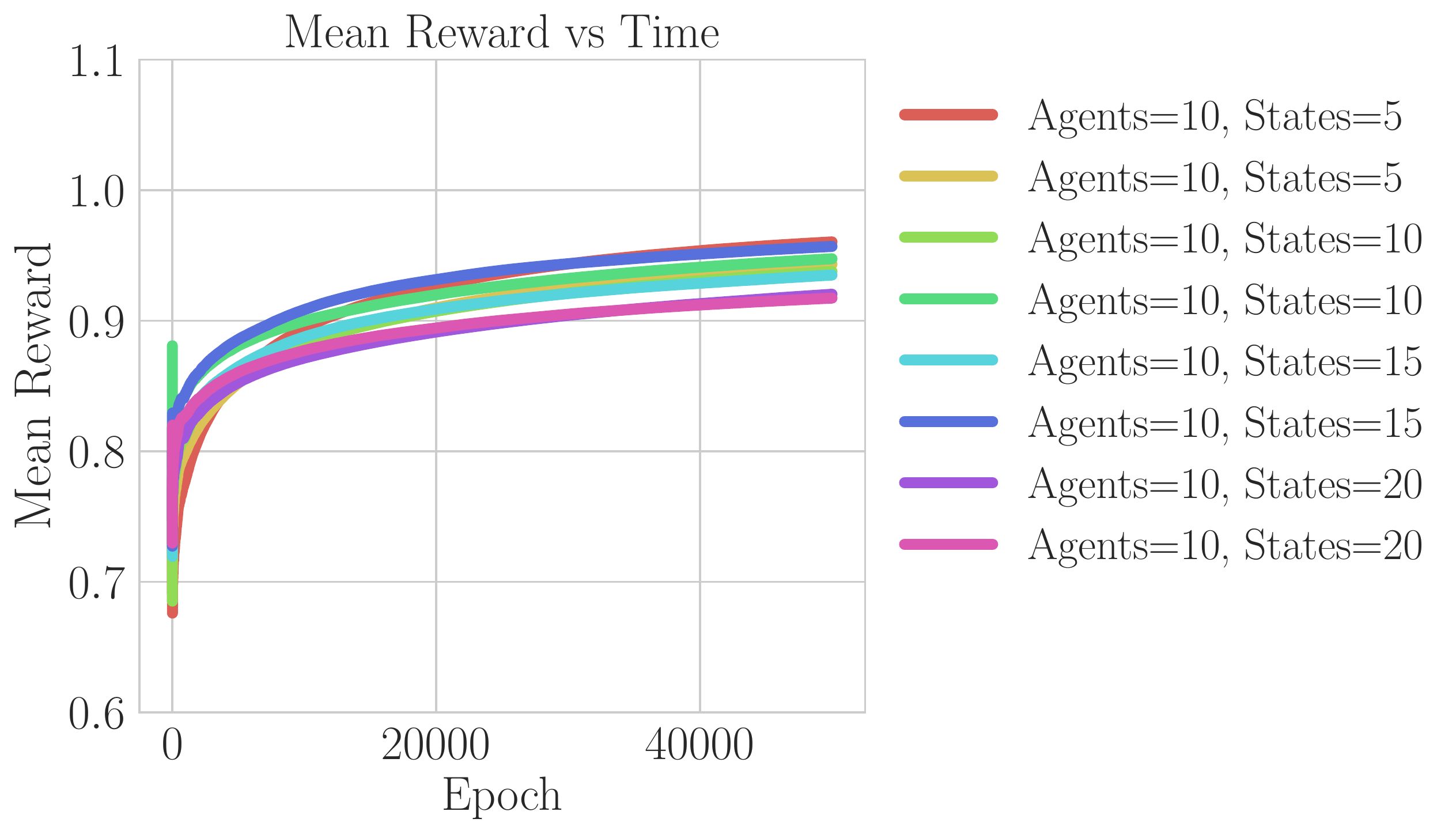}\label{fig:vary_states}}
    \subfloat[]{\includegraphics[width=.5\textwidth]{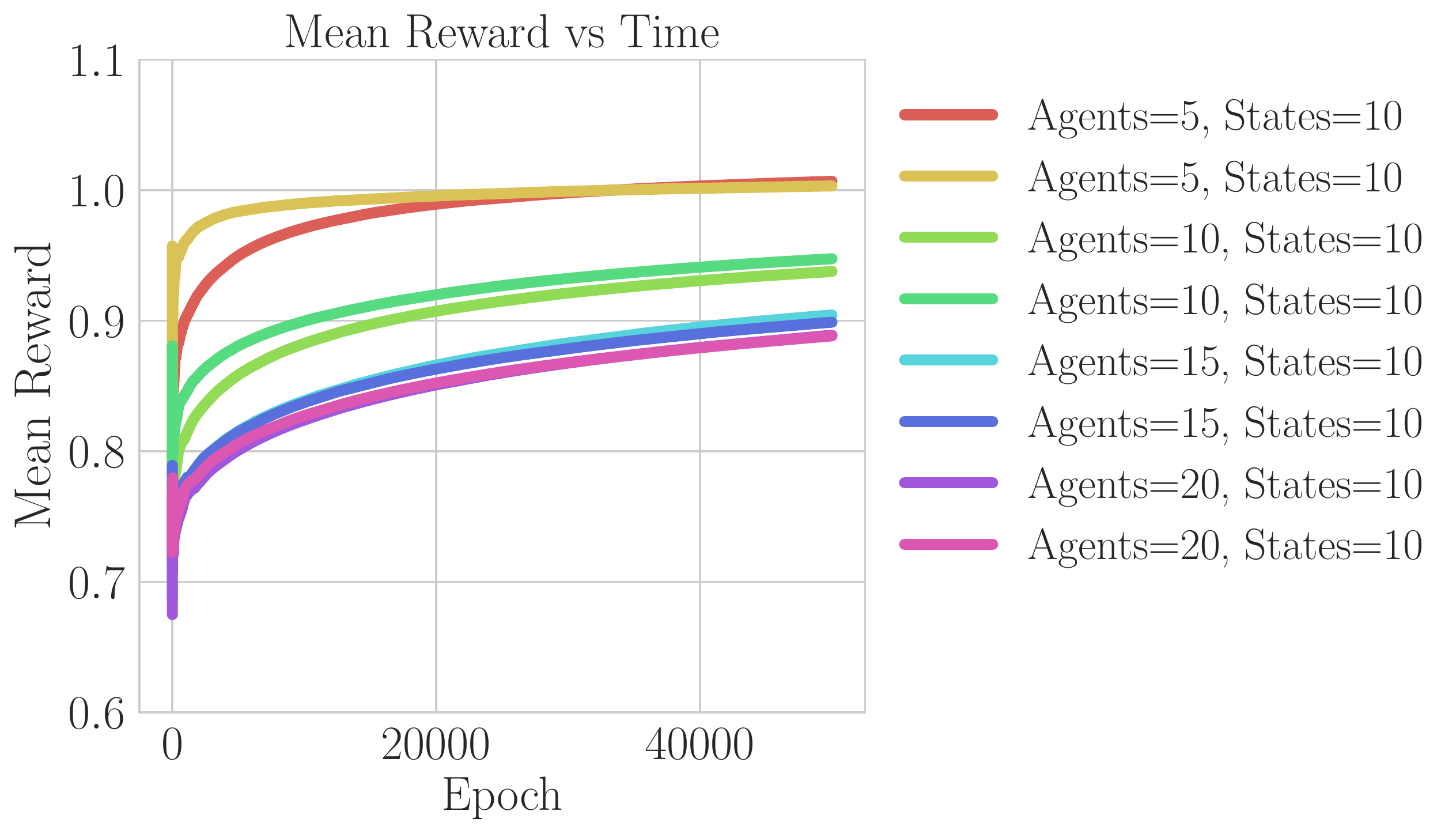}\label{fig:vary_agents}}
    \caption{Figure~\ref{fig:vary_states} presents results demonstrating how the performance of our algorithm varies with the number of states given that the number of agents and incentives is fixed for two instances of each configuration. Figure~\ref{fig:vary_agents} shows how the performance of the algorithm varies with the number of agents and incentives given that the number of states is fixed for two instances of each configuration.}
    \label{fig:vary_params}
\end{figure*}
\begin{figure*}[t]
    \subfloat[Static Demand]{\includegraphics[width=.4\textwidth]{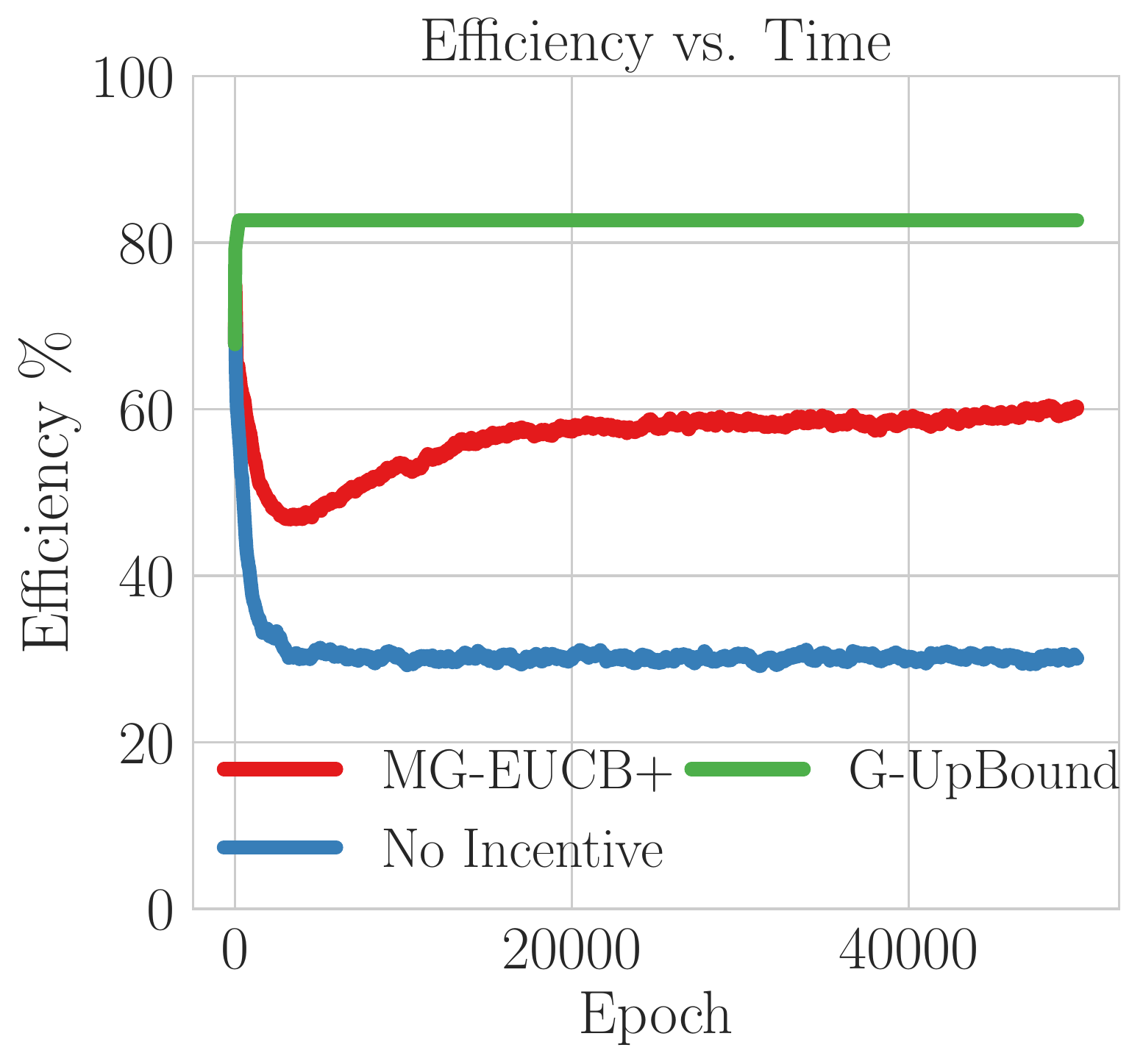}\label{fig:eff_static_utility}}
  \hfill  \subfloat[Random Demand]{\includegraphics[width=.4\textwidth]{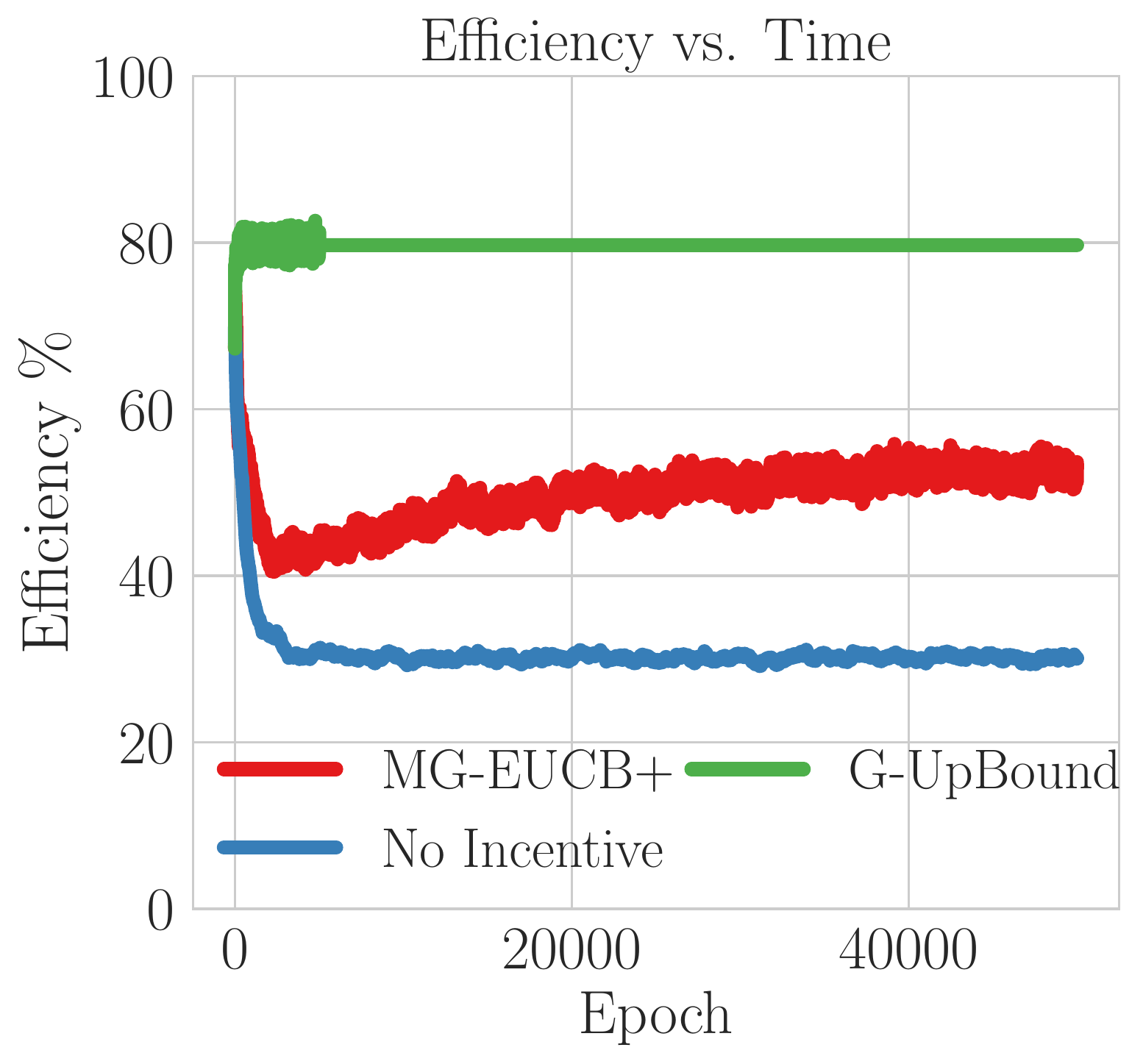}\label{fig:eff_random_utility}}
  \caption{Bike-share experiments with utility model: Figures~\ref{fig:eff_static_utility} and~\ref{fig:eff_random_utility} compare the efficiency of the bike-share system with two demand models and a utility based behavioral model under incentive matchings selected by MG-EUCB+ with upper and lower bounds given by the system performance when the incentive matching is given by computing the optimal greedy matching at each epoch based on the current state information and when no incentives are offered respectively.}
  \label{fig:utility}
\end{figure*}
\section{ADDITIONAL EXPERIMENTS}
\label{app:additional}
\subsection{COMPARISON OF TRADITIONAL UCB AND MG-EUCB FOR SIMPLE EXAMPLE}
\label{app:examplesim}

We return to the simple two-agent two-incentive instance depicted in Figure~\ref{fig:M1}. We ignore the capacity constraints by assuming that there is a single class $C_1$ such that every edge belongs to this class and $b_{C_1} = 2$. Clearly, this instance only admits two unique matchings $M^* = \{(a_1, i_1), (a_2, i_2)\}$---the optimum matching---and $M = \{(a_1, i_2), (a_2, i_1)\}$---the sub-optimal matching. 

As discussed previously, any traditional bandit approach that ignores the evolution of agent rewards would converge to the sub-optimal matching, i.e., $M$. To see why, observe that every time the algorithm selects the matching $M$ , both the
agents' states are reset to $\theta_1$ . Following this, when the algorithms `explores' the optimum matching, the reward
consistently happens to be zero since the agents are in state $\theta_1$. Owing to this, the traditional approach largely
underestimates the rewards for the (edges in the) optimum matching and converges to $M$. 

\begin{figure}[H]
    \includegraphics[scale=0.4]{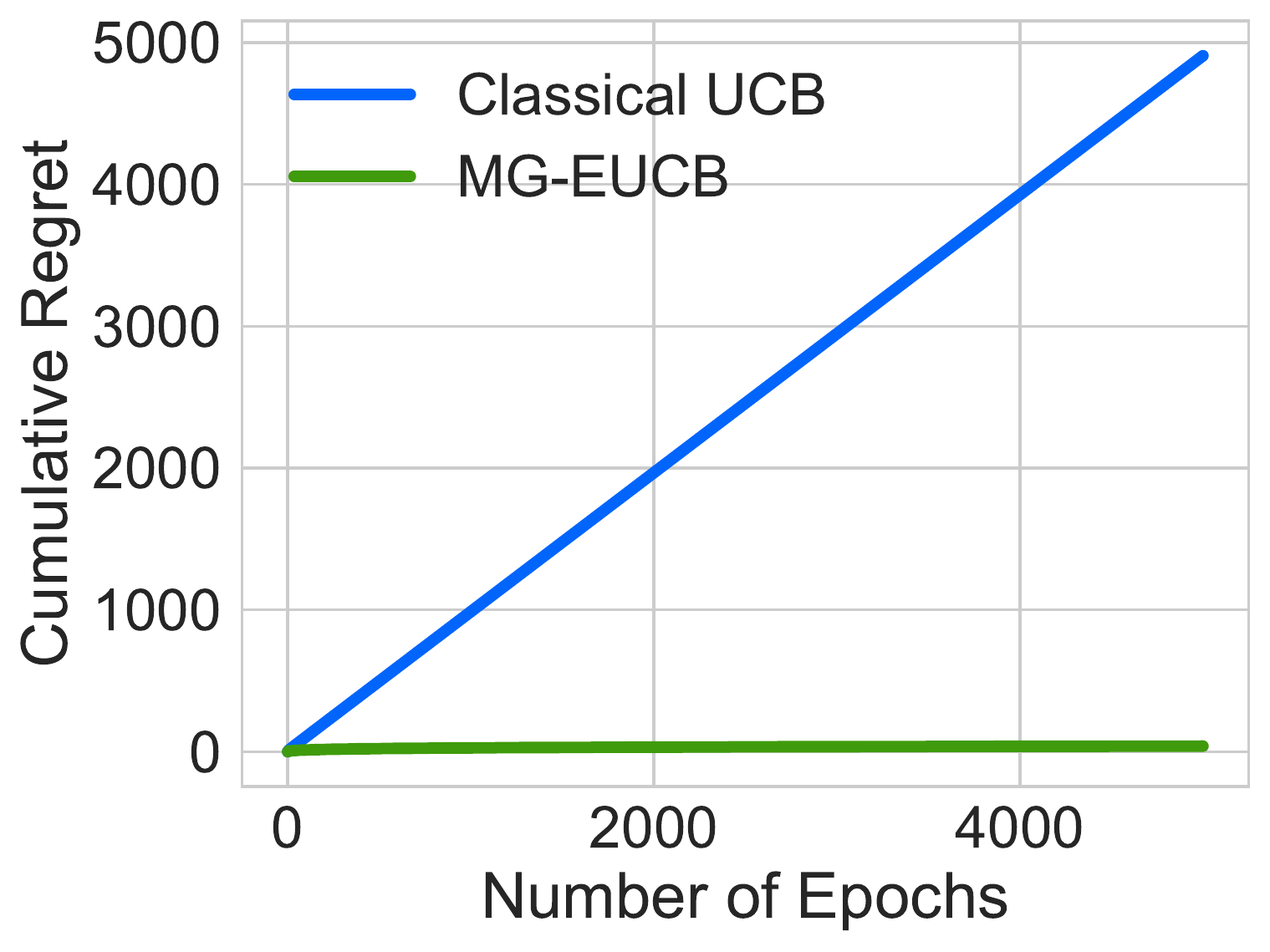}
    \caption{Comparison of the performance of classical UCB algorithms for matching problems versus the MatchGreedy-EpochUCB algorithm for the example depicted in Figure~\ref{fig:M1}.The length of horizon was $n=5000$.}
    \label{fig:M1sim}
\end{figure}

To validate this experimentally, we compare the performance of our MatchGreedy-EpochUCB algorithm described in Algorithm~\ref{alg:ucb} to a conventional implementation of the UCB algorithm for matching problems (e.g., as in~\cite{gai:2011aa,chenWYW16,kveton2015tight}). More specifically, we consider an implementation that runs for a total of $\sum_{i=1}^k \tau_k$ for some suitable set of parameters---in each iteration, the algorithm selects a matching based on the empirical rewards and the confidence bound. The iterations are then divided into rewards for convenience and the time-average reward in each epoch is computed and plotted alongside the same metric for the MG-EUCB algorithm in Figure~\ref{fig:M1sim}.

Our simulations support our prior conclusions. For example, after 5000 epochs, the classical UCB algorithm selects the sub-optimal matching over 99\% of the time. Owing to this reason, the classical algorithm has a regret that grows linearly with the length of the horizon whereas the regret of our algorithm is almost zero for this instance.

\subsection{ADDITIONAL SYNTHETIC EXPERIMENTS}
In our synthetic simulations we fixed the number of agents, incentives, and states equally as ${m = |\mathcal{A}| = |\mathcal{I}| = |\Theta_a| =10}$. We now present results in Figure~\ref{fig:vary_params} evaluating how the performance of our algorithm varies with each of these parameters. In Figure~\ref{fig:vary_states}, we observe that when the number of agents and incentives is fixed, the number of states has a negligible impact on the rate of convergence to the optimal solution. This indicates that within this range of states the Markov chains mix rapidly and the edge dependent constants in the regret bound do not significantly factor in. We find in Figure~\ref{fig:vary_agents}, as predicted by our regret bounds, the convergence slows as the number of agents in the problem increases.

\subsection{ADDITIONAL BIKE-SHARE DESCRIPTION AND EXPERIMENTS}
In this section we provide further motivation for the bike-sharing problem as a matching problem, more detail on our problem setup, as well as additional experimental results. Bike-share programs must deal with varying spatio-temporal demand to ensure that a high percentage of demand is met in order to satisfy customers and maximize profit. To avoid both pile-ups of bikes at popular destinations and depletion of bikes at stations with high demand, bike sharing companies manually replenish and manipulate the spatial supply of bikes. This is costly to companies and an alternative is to attempt to incentivize users to alter their paths in order to balance the spatial supply of bikes in such a way that meets future demand. A successful incentive system could reduce the need for manually replenishing the supply of bikes at stations, saving money and time as a result. 
\begin{figure}[H]
	\centering
	\includegraphics[width=.45\textwidth]{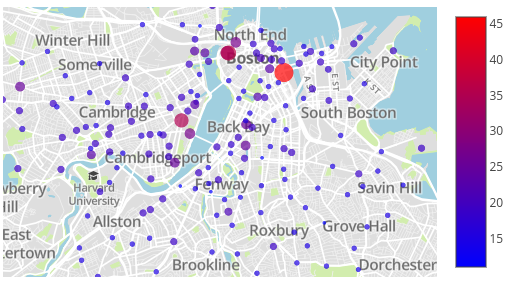}
	\caption{Heatmap of the scaled initial supply of the Boston Hubway stations. Each bubble indicates the location of a station and are scaled in size and colored according to the number of bikes available at the station.}
	\label{fig:map_supply}
\end{figure}
We consider the bike-share problem as a repeated game in our simulations. Specifically, at each epoch users move into the system seeking a bike from a station while simultaneously users transition from the location in which they picked up a bike to a location where they drop off the bike. In our simulations we allow the spatial supply of bikes to evolve based on the transitions of bikes between stations. We begin each simulation with the supply at each station given by the data scaled by a factor of two. As a result we have over $6000$ agents in the system that can move between close to $200$ stations.

We experimented with static and random demand models using quantities derived from the data. In the static demand model we set the demand between a directed pair of stations at each epoch to be the empirical mean of the number of transitions between the stations within $12$PM--$1$PM at each day over June, 2017 -- August, 2017. In our random demand model we used the empirical means as the parameter of a Poisson distribution from which we sampled the demand at each epoch for each directed pair of stations. To justify this choice we have included several representative probability mass functions for the demand between stations and the Poisson distributions that were fit to them in Figure~\ref{fig:dists}. We also applied goodness of fit tests to ensure this was a realistic modeling choice.
\begin{figure}[t]
	\centering
	\includegraphics[width=.45\textwidth]{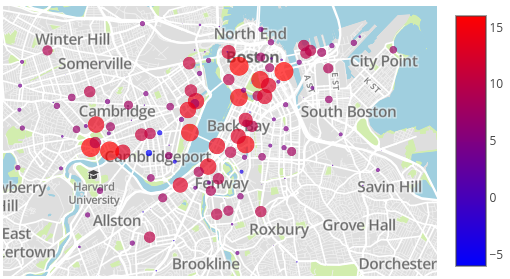}
	\label{fig:map_reject_change}
	\caption{This heatmap shows the spatial reduction in the number of rejections at each station in epoch 20000 from epoch 1000 corresponding to the result in Figure~\ref{fig:eff_static}. Positive numbers indicate how many fewer rejections occurred at the station at the later epoch than the earlier epoch. We observe a global reduction spatially in rejections nearly uniformly.}
\end{figure}

\begin{figure*}[t]
    \centering
    \subfloat[]{\includegraphics[width=.3\textwidth]{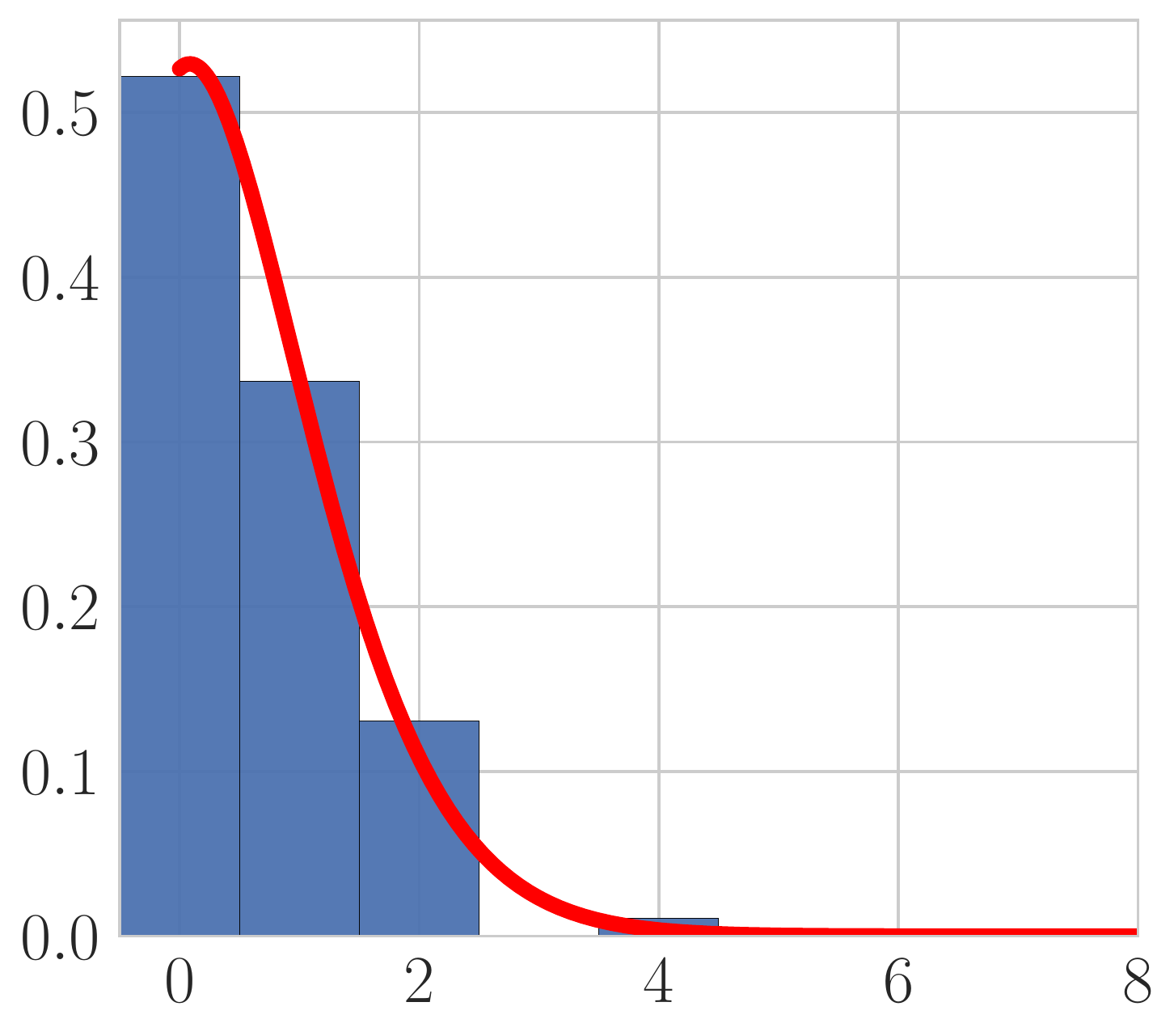}\label{fig:dist1}}
    \subfloat[]{\includegraphics[width=.3\textwidth]{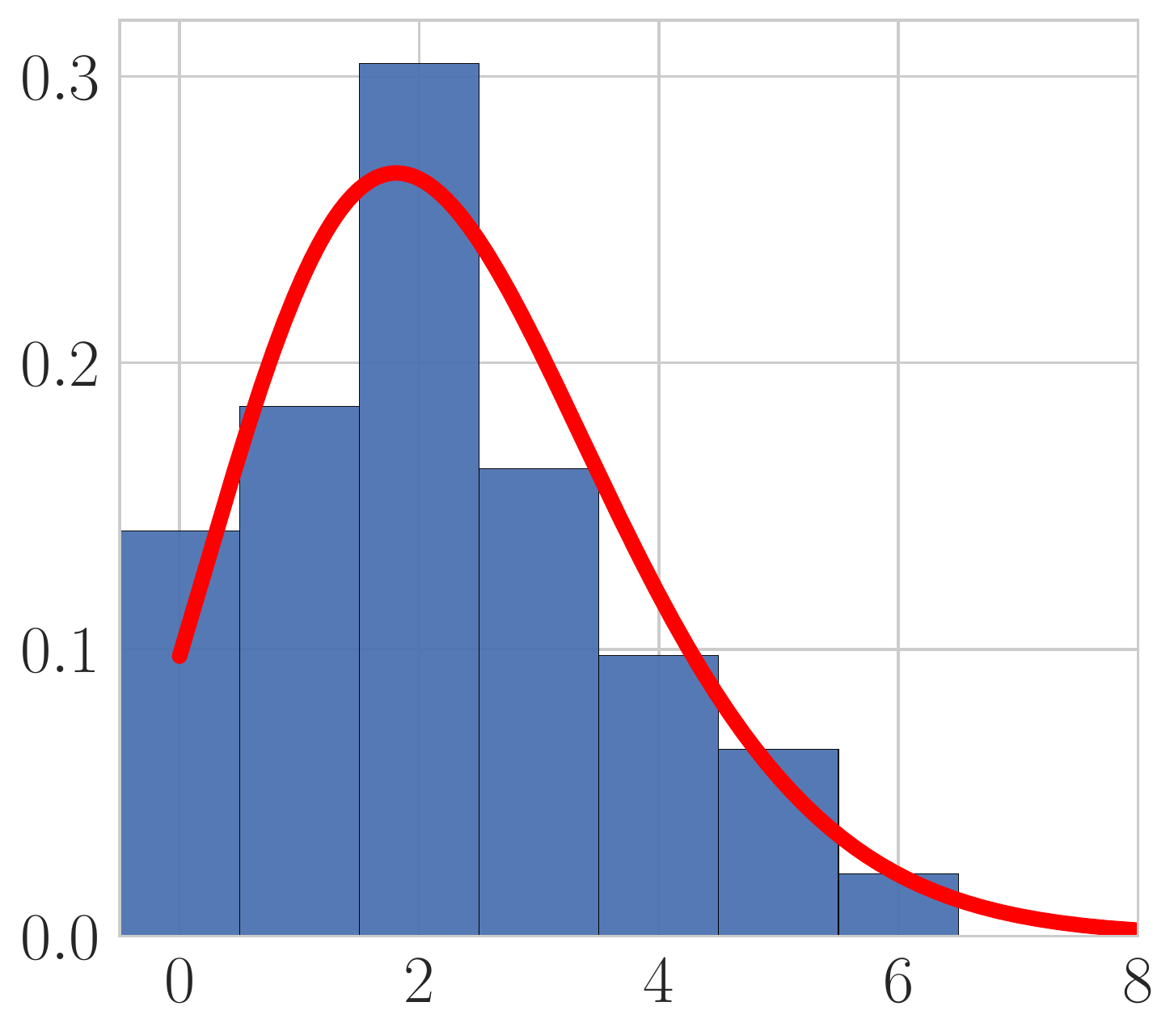}\label{fig:dist6}}
    \subfloat[]{\includegraphics[width=.3\textwidth]{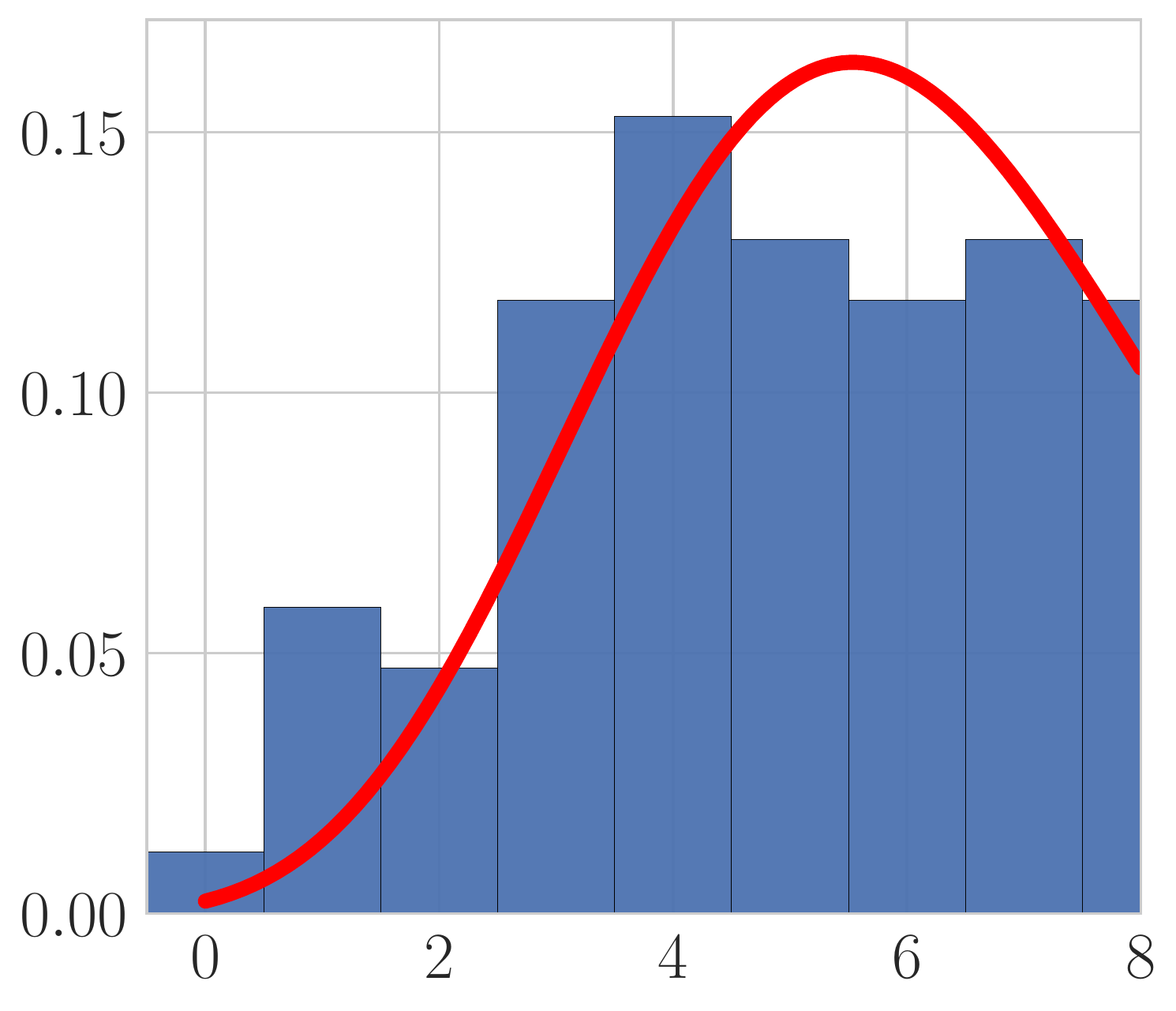}\label{fig:dist3}}
    \caption{Each empirical probability mass function in the figure gives the probability on the number of users that transitioned between a pair of stations in the Boston Hubway dataset between $12$PM--$1$PM each day between June, 2017 -- August, 2017. The red lines show the Poisson distribution that we fit to the distributions that we sampled from to generate random demand at each epoch of the simulation.}
    \label{fig:dists}
\end{figure*}

In our simulations we considered two behavioral models of the users in the system that govern how rewards are produced as well as the probability of a user accepting an incentive. As touched upon previously, in our bike-share model, associated with the state of a user are a distance threshold parameter and a parameter of a Bernouilli distribution. The distance threshold gives the maximum distance a user is willing to be re-routed and is drawn uniformly at random for each state in $[0, 4000]$ meters. The Bernouilli parameter gives the probability that a user will accept an incentive below its distance threshold for a particular state and is drawn uniformly at random in $[0, 1]$. In the primary behavioral model we consider based on a Bernouilli distribution presented in Figure~\ref{fig:bike_figs}, if the distance between the two stations of the proposed incentive is less than the threshold parameter associated with an agent's state the agent will accept the incentive with probability $p$ and give a reward of one, otherwise the incentive will be rejected and a reward of zero will be given. We also investigate a utility-based model; this model is the same as the Bernouilli based model with the slight modification that if an incentive is accepted following a successful realization of the Bernouilli draw, a reward is given that is proportional to the difference in distance between the threshold associated with a users state and the distance between the station the user intended to go to and the station of the proposed incentive.

We now give an overview of our results and the additional experiments we present in this section. We make two key favorable observations from the simulations in Figure~\ref{fig:bike_figs} in which we investigated static and random demand with the Bernoulli behavioral model. First, compared to a naive baseline of the convergence of the system without any incentives our algorithm is able to increase the efficiency of the system approximately $40$\% with the static demand model. Furthermore, the extension to random demand does not reduce the performance significantly. When comparing to an upper bound on performance we observe that our algorithm leads the system to approach this limit. 

The mean matching rewards presented in Figure~\ref{fig:mean_both} can be interpreted as the mean number of incentives that are accepted and equivalently the mean of users re-routed. This result indicates that on average less than $1$\% of users are matched to an incentive. This is a highly desirable property as it means we only need to influence a small part of the population in order to get significant performance gains. As a result, most users will only benefit from the incentive system, while from the planners perspective the minuscule cost of incentivizing only a small portion of the population is a beneficial.

We now show the results in Figure~\ref{fig:utility} of the static and random demand in combination with the utility based behavioral model. We generally draw the same conclusions as from Figure~\ref{fig:bike_figs} with somewhat lower performance for the system. This is an expected result as the users are more sensitive to the extra distance they must travel due to an incentive and they are therefore more difficult to incentivize. We note that we observed looking at the additional distances traveled due to an accepted incentive, that users under the utility based model do travel modestly less additional distance as a result of accepting an incentive than when we used the Bernouilli based model.

\section{IMPLEMENTATION DETAILS}
We make a small modification to the number of iterations within an epoch to
reduce computation time of the MG-EUCB algorithm. Specifically when the time-averaged reward has changed
by no more than $5\times 10^{-4}$ between consecutive iterations for $200$ iterations in a row---indicating the time averaged reward has converged---we end the epoch early. We find that this leads to the number of iterations in an epoch being roughly in the range of $1000$-$1500$. We observe this leads to a negligible change in the mean and cumulative rewards of the algorithm while significantly speeding up computation over a large horizon.

\section{Discussion}
In this work we developed a bandit algorithm for matching incentives to users, whose preferences are unknown a priori and evolving dynamically in time, in a resource constrained environment. We theoretically analyzed the problem and derived logarithmic gap-dependent regret bounds. There are several interesting future lines of work that we believe are worth pursuing. 

In this work, under the MDP dynamics we only investigated the combinatorial optimization problem of resource constrained matching and our proof techniques relied on the properties of the greedy matching paradigm. In future work, we are interested in attempting to extend this work to arbitrary combinatorial optimization problems with constraints in the case that the designer is allowed oracle access to solve the optimization problem, as has been done in the case without dynamics \cite{kveton2015tight, wen2015efficient}. 

The resource constraints that we considered were static over time. It is often the case that constraints of this form are time-varying or coupled over the decision-making horizon. A prominent example in online resource allocation is the Adwords problem. Due to the practical significance, we plan to explore if our model can be adapted to capture this richer class of constraints.

Finally, we would like to make our model increasingly realistic from the designer's and agents' perspectives. From the designer's point of view, this would be to incorporate incentive compatibility and fairness constraints. From the perspective of the agent, beyond the MDP dynamics, strategic behavior will be important to model and assess the impacts of going forward.  

\end{document}